%% file: AdaMomentum.tex
\documentclass[11pt]{article} %
\usepackage{url}
\usepackage{smile}
\usepackage{graphicx} %
\usepackage{algorithm}
\usepackage{algorithmic}
\usepackage{todonotes}
\usepackage{epstopdf}
\usepackage{wrapfig}
\usepackage[colorlinks, linkcolor=blue, anchorcolor=blue, citecolor=blue]{hyperref}
\usepackage[margin=1in]{geometry}
\usepackage[normalem]{ulem}
\usepackage[export]{adjustbox}%
\usepackage{mathtools, cuted}
\usepackage{natbib}
\usepackage{enumerate}
\usepackage{enumitem}

\usepackage{kpfonts}
\DeclareMathAlphabet{\mathsf}{OT1}{cmss}{m}{n}

\SetMathAlphabet{\mathsf}{bold}{OT1}{cmss}{bx}{n}

\providecommand{\norm}[1]{\|#1\|}

\begin{document}

\title{\huge \bf Rethinking Adam: A Twofold Exponential Moving Average Approach\thanks{The code will be made publicly available after the acceptance of the paper for publication.}}

\author
{Yizhou Wang$^\dag$, Yue Kang$^\ddag$, Can Qin$^\dag$, Huan Wang$^\dag$, \\ Yi Xu$^\dag$, Yulun Zhang$^\dag$, and Yun Fu$^\dag$ \\
$^\dag$ Northeastern University, $^\ddag$ University of California, Davis \\
\texttt{wyzjack990122@gmail.com}, \texttt{yuekang@ucdavis.edu}, \texttt{\{qin.ca,xu.yi\}@northeastern.edu}, \\ \texttt{huan.wang.cool@gmail.com}, \texttt{yulun100@gmail.com}, \texttt{yunfu@ece.neu.edu}
}

\date{}

\maketitle

\begin{abstract}
  Adaptive gradient methods, e.g. \textsc{Adam}, have achieved tremendous success in machine learning. Scaling the learning rate element-wisely by a certain form of second moment estimate of gradients, such methods are able to attain rapid training of modern deep neural networks. Nevertheless, they are observed to suffer from compromised generalization ability compared with stochastic gradient descent (\textsc{SGD}) and tend to be trapped in local minima at an early stage during training. Intriguingly, we discover that substituting the gradient in the second raw moment estimate term with its momentumized version in \textsc{Adam} can resolve the issue. The intuition is that gradient with momentum contains more accurate directional information and therefore its second moment estimation is a more favorable option for learning rate scaling than that of the raw gradient. Thereby we propose \textsc{AdaMomentum} as a new optimizer reaching the goal of training fast while generalizing much better. We further develop a theory to back up the improvement in generalization and provide convergence guarantees under both convex and nonconvex settings. Extensive experiments on a wide range of tasks and models demonstrate that \textsc{AdaMomentum} exhibits state-of-the-art performance and superior training stability consistently. %
\end{abstract}

\input{intro.tex}

\input{method.tex}

\input{why_adamomentum_optimization.tex}

\input{why_adamomentum_generalization_new.tex}

\input{convergence_convex.tex}

\input{convergence_nonconvex_new.tex}

\input{experiment.tex}

\input{conclusion.tex}

\bibliographystyle{ims}
\bibliography{capacity}

\newpage
\appendix
\input{appendix_escapeminima.tex}

\input{appendix_convex.tex}
\input{appendix_nonconvex.tex}

\input{appendix_exp.tex}

\end{document}

%% file: intro.tex
\section{Introduction}
Prevailing first-order optimization algorithms in modern machine learning can be classified into two categories. One is stochastic gradient descent (SGD)~\citep{robbins1951stochastic}, which is widely adopted due to its low memory cost and outstanding performance. SGDM~\citep{sutskever2013importance} which incorporates the notion of momentum into SGD, has become the best choice for optimizer in computer vision. The drawback of SGD(M) is that it scales the gradient uniformly in all directions, making the training slow especially at the begining and fail to optimize complicated models well beyond Convolutional Neural Networks (CNN). The other type is adaptive gradient methods. Unlike SGD, adaptive gradient optimizers adapt the stepsize (a.k.a. learning rate) elementwise according to the gradient values. Specifically, they scale the gradient by the square roots of some form of the running average of the squared values of the past gradients. Popular examples include AdaGrad~\citep{duchi2011adaptive}, RMSprop~\citep{hinton2012neural} and Adam~\citep{DBLP:journals/corr/KingmaB14} etc. Adam, in particular, has become the default choice for many machine learning application areas owing to its rapid speed and outstanding ability to handle sophisticated loss curvatures. 

Despite their fast speed in the early training phase, adaptive gradient methods are found by studies~\citep{wilson2017marginal,zhou2020generalizationdeep} to be more likely to exhibit poorer generalization ability than SGD. This is discouraging because the ultimate goal of training in many machine learning tasks is to exhibit favorable performance during testing phase. In recent years researchers have put many efforts to mitigate the deficiencies of adaptive gradient algorithms. AMSGrad~\citep{reddi2018convergence} corrects the errors in the convergence analysis of Adam and proposes a faster version. Yogi~\citep{reddi2018adaptive} takes the effect of batch size into consideration. M-SVAG~\citep{balles2018dissecting} transfers the variance adaptation mechanism from Adam to SGD. AdamW~\citep{loshchilov2017decoupled} first-time decouples weight decay from gradient descent for Adam-alike algorithms. SWATS~\citep{keskar2017improving} switches from Adam to SGD throughout the training process via a hard schedule and AdaBound~\citep{luo2019adaptive} switches with a smooth transation by imposing dynamic bounds on stepsizes. RAdam~\citep{liu2019variance} rectifies the variance of the adaptive learning rate through investigating the theory behind warmup heuristic~\citep{vaswani2017attention,popel2018training}. AdaBelief~\citep{zhuang2020adabelief} adapts stepsizes by the belief in the observed gradients. Nevertheless, most of the above variants can only surpass (as they claim) Adam or SGD in limited tasks or under specifically and carefully defined scenarios. Till today, SGD and Adam are still the top options in machine learning, especially deep learning~\citep{schmidt2020descending}. Conventional rules for choosing optimizers are: from task perspective, choose SGDM for vision, and Adam (or AdamW) for language and speech; from model perspective, choose SGDM for Fully Connected Networks and CNNs, and Adam for Recurrent Neural Networks (RNN)~\citep{cho-etal-2014-properties,hochreiter1997long}, Transformers~\citep{vaswani2017attention} and Generative Adversarual Networks (GAN)~\citep{goodfellow2014generative}.
Based on the above observations, a natural question is:

\textit{Is there an efficient adaptive gradient algorithm that can converge fast and meanwhile generalize well?}

In this work, we are delighted to discover that simply replacing the gradient term in the second moment estimation term of Adam with its momentumized version can achieve this goal. Our idea comes from the origin of Adam optimizer, which is a combination of RMSprop and SGDM. RMSprop scales the current gradient by the square root of the exponential moving average (EMA) of the squared past gradients, and Adam replaces the raw gradient in the numerator of the update term of RMSprop with its EMA form, i.e., with momentum. Since the EMA of gradient is a more accurate estimation of the appropriate direction to descent, we consider putting it in the second moment estimation term as well. We find such operation makes the optimizer more suitable for the general loss curvature and can theoretically converge to minima that generalize better. Extensive experiments on a broad range of tasks and models indicate that: without bells and whistles, our proposed optimizer can be as good as SGDM on vision problems and outperforms all the competitor optimizers in other tasks, meanwhile maintaining fast convergence speed. Our algorithm is efficient with no additional memory cost, and applicable to a wide range of scenarios in machine learning. More importantly, AdaMomentum requires little effort in hyperparameter tuning and the default parameter setting for adaptive gradient method works well consistently in our algorithm. 

\paragraph{Notation} We use $t,T$ to symbolize the current and total iteration number in the optimization process. $\theta \in \RR^d$ denotes the model parameter and $f(\theta) \in \RR$ denotes the loss function. We further use $\theta_t$ to denote the parameter at step $t$ and $f_t$ to denote the noisy realization of $f$ at time $t$ because of the mini-batch stochastic gradient mechanism. $g_t$ denotes the $t$-th time gradient and $\alpha$ denotes stepsize. $m_t, v_t$ represent the EMA of the gradient and the second moment estimation term at time $t$ of adaptive gradient methods respectively. $\epsilon$ is a small constant number added in adaptive gradient methods to refrain the denominator from being too close to zero. $\beta_1, \beta_2$ are the decaying parameter in the EMA formulation of $m_t$ and $v_t$ correspondingly. For any vectors $a,b \in \RR^d$, we employ $\sqrt{a}, a^2, |a|, a/b, a\ge b, a\le b$ for elementwise square root, square, absolute value, division, greater or equal to, less than or equal to respectively. For any $1 \le i \le d$, $\theta_{t,i}$ denotes the $i$-th element of $\theta_t$. Given a vector $x \in \RR^d$, we use $\norm{x}_2 $ to denote its $l_2$-norm and $\norm{x}_{\infty}$ to denote its $l_\infty$-norm.

%% file: method.tex
\section{Algorithm}

\paragraph{Preliminaries \& Motivation}

\begin{table}[t]
  \caption{Comparison of AdaMomentum and classic adaptive gradient methods in $m_t$ and $v_t$ in~\eqref{eq: generic-form}.}
  \label{table: comparison-nd}
  \centering
  \scalebox{0.87}{
  \begin{tabular}{lcc}
  \toprule[1pt]
   Optimizer & $m_t$ & $v_t$ \\ \midrule
   SGD & $g_t$ & $1$ \\ 
   Rprop & $g_t$ & $g_t^2$ \\ 
   RMSprop & $g_t$ & $(1-\beta_2)\sum_{i=1}^t \beta_2^{t-i}g_i^2$\\ 
   Adam &$(1-\beta_1)\sum_{i=1}^t \beta_1^{t-i}g_i$ & $(1-\beta_2)\sum_{i=1}^t \beta_2^{t-i}g_i^2$ \\ \midrule
   {\bf Ours} & $(1-\beta_1)\sum_{i=1}^t \beta_1^{t-i}g_i$ & $(1-\beta_2)\sum_{i=1}^t \beta_2^{t-i}{\bbm_i}^2$\\
   \bottomrule[1pt]
  \end{tabular}
  }
\end{table}
Omitting the debiasing operation and the damping term $\epsilon$, the adaptive gradient methods can be generally written in the following form:
\small
\begin{equation}\label{eq: generic-form}
  \theta_{t+1} = \theta_t - \alpha \frac{m_t}{\sqrt{v_t}}.
\end{equation}
\normalsize
Here $m_t, v_t$ are called the first and second moment estimation terms. When $m_t=g_t$ and $v_t=1$, \eqref{eq: generic-form} degenerates to the vanilla SGD. Rprop~\citep{duchi2011adaptive} is the pioneering work using the notion of adaptive learning rate, in which $m_t=g_t$ and $v_t =g_t^2$. Actually it is equivalent to only using the sign of gradients for different weight parameters. RMSprop~\citep{hinton2012neural} forces the number divided to be similar for adjacent mini-batches by incorporating momentum acceleration into $v_t$. Adam~\citep{DBLP:journals/corr/KingmaB14} is built upon RMSprop in which it turns $g_t$ into momentumized version. Both RMSprop and Adam boost their performance thanks to the smoothing property of EMA. Due to the fact that the EMA of gradient is a more accurate estimation than raw gradient, we deem that there is no reason to use $g_t$ in lieu of $m_t$ in second moment estimation term $v_t$. Therefore we propose to replace the $g_i$s in $v_t$ of Adam with its EMA version $m_i$s, which further smooths the EMA. Hence our $v_t$ employs a twofold EMA approach, i.e., the EMA of the square of the EMA of the past gradients.

\begin{algorithm}[tb]
  \caption{AdaMomentum (ours). All mathematical operations are element-wise.}
  \label{alg: adamomentum}
  \begin{algorithmic}
    \STATE {\bfseries Initialization} : Parameter initialization $\theta_0$, step size $\alpha$, damping term $\epsilon$, $m_0 \leftarrow 0, v_0 \leftarrow 0, t\leftarrow 0$
    \WHILE{$\theta_t$ not converged}
        \STATE $t \leftarrow t+1 $ %
        \STATE $g_t \leftarrow \nabla_{\theta}f_t(\theta_{t-1}) $ %
        \STATE $m_t \leftarrow \beta_1 m_{t-1} + (1-\beta_1) g_t $ %
        \STATE $v_t \leftarrow \beta_2 v_{t-1} + (1-\beta_2) {\color{red} m_t}^2 + {\color{blue} \epsilon} $ %
        \STATE $\hat{m_t} \leftarrow m_t/(1-\beta_1^t) $ %
        \STATE $\hat{v_t} \leftarrow v_t/(1-\beta_2^t) $ %
        \STATE $\theta_t \leftarrow \theta_{t-1} - \alpha \cdot \hat{m}_t/\sqrt{\hat{v}_t} $ %
    \ENDWHILE
  \end{algorithmic}
\end{algorithm}

\paragraph{Detailed Algorithm} The detailed procedure of our proposed optimizer is displayed in Algorithm~\ref{alg: adamomentum}. There are two major modifications based on Adam, which are marked in red and blue respectively. One is that we replace the $g_t$ in $v_t$ of Adam with $m_t$, which is the momentumized gradient. Hence we name our proposed optimzier as \textit{AdaMomentum}. The other is the location of $\epsilon$ (in Adam $\epsilon$ is added after $\sqrt{\cdot}$ in line 10 of Alg.\ref{alg: adamomentum}). We discover that moving the adding of $\epsilon$ term from outside the radical symbol to inside can consistently enhance performance. To the best of our knowledge, our method is the first attempt to put momentumized gradient in the second moment estimation term of adaptive gradient methods. Note that although the modifications seem simple to some degree, they can lead to siginificant changes in the performance of an adaptive gradient optimizer due to the iterative nature of optimization methods, which will also be elaborated in the following sections.

%% file: why_adamomentum_optimization.tex
\begin{figure*}[tb]
  \centering
  \includegraphics[width=0.95\linewidth]{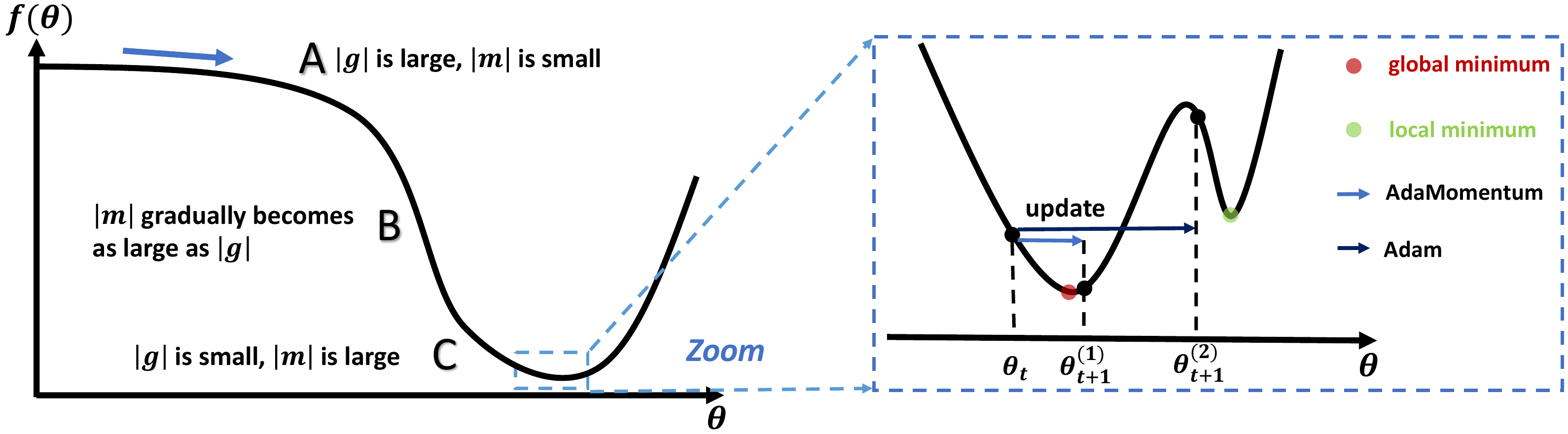}
  \vspace{-2mm}
  \caption{Illustration of the optimization process of Adam and AdaMomentum. A general loss curve can be composed to three areas: {\bf A)} transition from a plateau to a downgrade; {\bf B)} a steep downgrade; {\bf C)} from downgrade to entering the basin containing the optimum.  An ideal optimizer is expected to sustain large stepsize before reaching the optimum, and reduce its stepsize near the optimum. Compared to Adam, AdaMomentum can adapt the effective stepsize more appropriately along the loss curve, and maintain smaller stepsize near the convergence, which contributes to stable training and better convergence. Please refer to Section~\ref{sec: curvation} for more detailed analysis.}
  \label{fig: curvature}
\end{figure*}

\section{Why AdaMomentum over Adam?}\label{sec: why_adamm}
\subsection{AdaMomentum is More Suitable for General Loss Curvature}\label{sec: curvation}

In this section, we show that AdaMomentum can converge to (global) minima faster than Adam does via illustration. The left part of Figure~\ref{fig: curvature} is the process of optimization from a plateau to a basin area, where a global optimum is assumed to exist. The right part is the zoomed-in version of the situation near the minimum, where we have some peaks and valleys. This phenomenon frequently takes place in optimization since there is only one global minimum with probably a great number of local minima surrounding~\citep{hochreiter1997flat,keskar2016large}. 

\paragraph{Benefits of Substituting $g_t$ with $m_t$.} We first explain how substituting $m_t$ for $g_t$ in the preconditioner $v_t$ can improve training via decomposing the trajectory of parameter point along the loss curve. {\bf 1)} In area A, the parameter point starts to slide down the curve and $|g_t|$ begins to enlarge abruptly. So the actual stepsize $\alpha/\sqrt{v_t}$ is small for Adam. However the absolute value of the momentumized gradient $m_t$ is small since it is the EMA of the past gradients, making $\alpha/\sqrt{v_t}$ still large for AdaMomentum. Hence AdaMomentum can maintain higher training speed than Adam in this changing corner of the loss curve, which is what an optimal optimizer should do. {\bf 2)} In area B, since the exponential moving average decays the impact of past gradients exponentially w.r.t. $t$, the magnitude of the elements of $m_t$ will gradually becomes as large as $g_t$. {\bf 3)} In area C, when the parameter approaches the basin, the magnitude of $g_t$ decreases, making the stepsizes of Adam increase immediately. In contrast, the stepsize of AdaMomentum is still comparatively small as $|m_t|$ is still much larger than $|g_t|$, which is desired for an ideal optimizer. 
Small stepsize near optimum has benefits for convergence and stability. A more concrete illustration is given in the right part of Figure~\ref{fig: curvature}. If the stepsize is too large (e.g. in Adam), the weight parameter $\theta_t$ may rush to $\theta_{t+1}^{(2)}$  and miss the global optimum. In contrast, small stepsize can guarantee the parameter to be close to the global minimum (see $\theta_{t+1}^{(1)}$) even if there may be tiny oscillations within the basin before the final convergence.

\paragraph{Benefits of Changing the Location of $\epsilon$.} Next we elaborate why putting $\epsilon$ under the $\sqrt{\cdot}$ is beneficial. We denote the debiased second moment estimation in AdaMomentum as $\hat{v}_t$ and the second moment estimation term without $\epsilon$ as $\hat{v}_t'$. By simple calculation, we have
\small
\begin{align*}
  \hat{v}_t &= \left((1-\beta_2)/(1-\beta_2^t) \right)\cdot \sum_{i=1}^t \beta_2^{t-i} m_i^2 + \frac{\epsilon}{1-\beta_2},\\
  \hat{v}_t' &= \left((1-\beta_2)/(1-\beta_2^t)\right)\cdot \sum_{i=1}^t \beta_2^{t-i} m_i^2.
\end{align*}
\normalsize
Hence we have $\hat{v}_t = \hat{v}_t' + \epsilon/(1-\beta_2)$. Then the actual stepsizes are $\alpha/(\sqrt{{\hat{v}_t'} + \epsilon/(1-\beta_2)})$ and $\alpha/(\sqrt{\hat{v}_t'} + \epsilon)$ respectively. In the final stage of optimization, $\hat{v}_t'$ is very close to $0$ (because the values of gradients are near $0$) and far less than $\epsilon$ hence the actual stepsizes can be approximately written as $\sqrt{1-\beta_2}\alpha/\sqrt{\epsilon}$ and $\alpha/\epsilon$. As $\epsilon$ usually takes very tiny values ranging from $10^{-8}$ to $10^{-16}$ and $\beta_2$ usually take values that are extremely close to $1$ (usually $0.999$), we have $\sqrt{1-\beta_2}\alpha/\sqrt{\epsilon} \ll \alpha/\epsilon$. Therefore we may reasonably come to the conclusion that after moving $\epsilon$ term into the radical symbol, AdaMomentum further reduces the stepsizes when the training is near minima, which contributes to enhancing convergence and stability as we have discussed above.

%% file: why_adamomentum_generalization_new.tex
\subsection{AdaMomentum Converges to Minima that Generalize Better} \label{sec: escape_minima}

The outline of Adam and our proposed AdaMomentum can be written in the following unified form:
\small
\begin{align}
    m_t = \beta_1 &m_{t-1} + (1-\beta_1) g_t, \quad v_t = \beta_2 v_{t-1} + (1-\beta_2) k_t^2, \nonumber \\
    &\theta_{t+1} = \theta_t - \alpha \, m_t \Big{/} \left((1-\beta_1^t)\sqrt{v_t/(1-\beta_2^t)}\right). \label{adam_alike_eq}
\end{align}
\normalsize

where $k_t = g_t$ in Adam and $k_t = m_t$ in AdaMomentum. Inspired by a line of work~\citep{pavlyukevich2011first,simsekli2019tail,zhou2020generalizationdeep}, we can consider~\eqref{adam_alike_eq} as a discretization of a continuous-time process and reformulate it as its corresponding L\'evy-driven stochastic differential equation (SDE). Assuming that the gradient noise $\zeta_t = g_t - \nabla f(\theta_t)$ is centered symmetric $\tilde\alpha$-stable ($\mathcal{S}\tilde\alpha\mathcal{S}$)~\citep{levy1954theorie} distributed with covariance matrix $\Sigma_t$ possessing a heavy-tailed signature ($\tilde\alpha \in (0,2]$), then we are able to derive the  L\'evy-driven SDE of~\eqref{adam_alike_eq} as:
\small
\begin{align}
    &d\theta_t = -q_t R_t^{-1} m_t dt+\upsilon R_t^{-1} \Sigma_t d L_t, \\
     &dm_t = \beta_1(\nabla f(\theta_t) - m_t), \; dv_t = \beta_2(k_t^2 - v_t), \label{sde_eq}
\end{align}
\normalsize
where $R_t = \text{diag}(\sqrt{v_t/(1-\beta_2^t)}), \upsilon = \alpha^{1-1/\tilde\alpha}, q_t = 1/(1-\beta_1^t)$ and $L_t$ is the $\tilde \alpha$-stable L\'evy motion with independent components. We are interested in the local stability of the optimizers and therefore we suppose process~\eqref{sde_eq} is initialized in a local basin $\boldsymbol{\Omega}$ with a minimum $\theta^*$ (w.l.o.g., we assume $\theta^*={\bf 0}$). To investigate the escaping behavior of $\theta_t$, we first introduce two technical definitions.
\begin{definition}[{\bf Radon Measure}~\citep{simon1983lectures}]
    If a measure $m(\cdot)$ defined on the $\sigma$-algebra of Borel sets of a Hausdorff topological space $X$ is 1) inner regular on open sets, 2) outer regular on all Borel sets, and 3) finite on all compact sets, then the measure is called a Radon measure.
\end{definition}
\begin{definition}[{\bf Escaping Time \& Escaping Set}]\label{def: escaping}
    We define escaping time $ \Gamma \coloneqq \inf\{t \geq 0 \, : \, \theta_t \not\in \boldsymbol{\Omega}^{-\upsilon^\gamma}\}, \ \text{where }  \boldsymbol{\Omega}^{-\upsilon^\gamma} = \{y \in \boldsymbol{\Omega} \, : \, \text{dis}(\partial \boldsymbol{\Omega}, y ) \geq \upsilon^\gamma\}$. Here $\gamma>0$ is a constant. We define escaping set  $\Upsilon \coloneqq \{y \in \mathbb{R}^d \, : \, R_{\theta^*}^{-1} \Sigma_{\theta^*} y \not\in \boldsymbol{\Omega}^{-\upsilon^\gamma}\}$, where $\Sigma_{\theta^*} = \lim_{\theta_t \to \theta^*} \Sigma_{t}, R_{\theta^*} = \lim_{\theta_t \to \theta^*} R_{t}$.
\end{definition}
We study the relationship between $\Gamma$ and $\Upsilon$ and impose some standard assumptions before proceeding.

\begin{assumption}
\label{assu:escape1}
$f$ is non-negative with an upper bound, and locally $\mu$-strongly convex in $\boldsymbol{\Omega}$. 
\end{assumption}
\begin{assumption}\label{assu:escape2}
There exists some constant $L>0$, s.t. $\norm{\nabla f(x) - \nabla f(y)}_2 \leq L \norm{x-y}_2, \forall x,y$.
\end{assumption}
\begin{assumption}\label{assu:escape3}
    We assume that $\int_0^{\Gamma} \langle \nabla f(\theta_t)/(1 + f(\theta_t)), q_t R_t^{-1} m_t \rangle \, dt \geq 0$ a.e., and  $\beta_1 \leq \beta_2 \leq 2 \beta_1$. We further suppose that there exist $v_-, v_+>0$ s.t. each coordinate of $\sqrt{v_t}$ can be uniformly bounded in $(v_-,v_+)$ and there exist $\tau_m,\tau>0$ s.t. $\norm{ m_t - \hat{m}_t }_2 \leq \tau_m \norm{ \int_0^{t-} (m_x - \hat{m}_x) \, dx }_2$ and $\norm{ \hat{m}_t }_2 \geq \tau \norm{ \nabla f(\hat{\theta}_t)}_2$, where $\hat{m}_t$ and $\hat{\theta}_t$ are calculated by solving equation~\eqref{sde_eq} with $\upsilon = 0$. 
\end{assumption}
Assumption~\ref{assu:escape1} and~\ref{assu:escape2} impose some standard assumptions of stochastic optimization~\cite{ghadimi2013stochastic,johnson2013accelerating}. Assumption~\ref{assu:escape3} requires momentumized gradient $m_t$ and $\nabla f(\theta_t)$ to have similar directions for most of the time, which have been empirically justified to be true in Adam~\citep{zhou2020generalizationdeep}.
Based on the above assumptions, we can prove that for algorithm of form~\eqref{adam_alike_eq}, the expected escaping time is inversely proportional to the Radon measure of the escaping set:
\begin{lemma}\label{thm: escape_time}
Under Assumptions~\ref{assu:escape1}-\ref{assu:escape3}, let $\upsilon^{\tilde \alpha + 1} = \Theta(\tilde \alpha)$ and $\ln{(2\Delta/(\mu \upsilon^{1/3}))} \leq 2 \mu \tau(\beta_1 - \beta_2/4)/(\beta_1 v_+ + \mu \tau)$, where $\Delta = f(\theta_0) - f(\theta^*)$. Then given any $\theta_0 \in \boldsymbol{\Omega}^{-2\upsilon^\gamma}$, for~\eqref{sde_eq} we have
\small
\begin{equation*}
    \mathbb{E}(\Gamma) = {\Theta}(\upsilon/m(\Upsilon)),
\end{equation*}
\normalsize
where $m(\cdot)$ is a non-zero Radon measure satisfying that $m(\mathcal{U}) < m(\mathcal{V})$ if $\mathcal{U} \subset \mathcal{V}$.
\end{lemma}

Because larger set has larger volume, i.e., $   V(\mathcal{U}) \leq V(\mathcal{V})$ if $\mathcal{U} \subset \mathcal{V}$, from Lemma~\ref{thm: escape_time} we have the escaping time is negatively correlated with the volume of the set $\Upsilon$. Therefore, we can come to the conclusion that for both Adam and AdaMomentum, if the basin $\boldsymbol{\Omega}$ is sharp which is ubiquitous during the early stage of training, $\Upsilon$ has a large Radon measure, which leads to smaller escaping time $\Gamma$. This means both Adam and AdaMomentum prefer relatively flat or asymmetric basin~\cite{he2019asymmetric} through the training. 

On the other hand, upon encountering a comparatively flat basin or asymmetric valley $\boldsymbol{\Omega}$, we are able to prove that AdaMomentum will stay longer inside. Before we proceed, we need to impose two mild assumptions.
\begin{assumption}\label{asp: gradient-unbiased-noise-ind} 
 The $l_\infty$ norm of $\nabla f$ is upper bounded by some constant G, \textit{i.e.} $\norm{\nabla f(x)}_\infty \leq G, \forall x$. 
\end{assumption}

\begin{assumption}\label{asp:noise_decay}
    For AdaMomentum, there exists $T_0 \in \mathbb{N}$ s.t., $\mathbb{E}(\zeta_t^2) \leq \beta_1\mathbb{E}(m_{t-1}^2)/(2-\beta_1)$ when $t>T_0$. 
\end{assumption}

\begin{figure}[tb]
    \centering
    \includegraphics[width=0.75\linewidth]{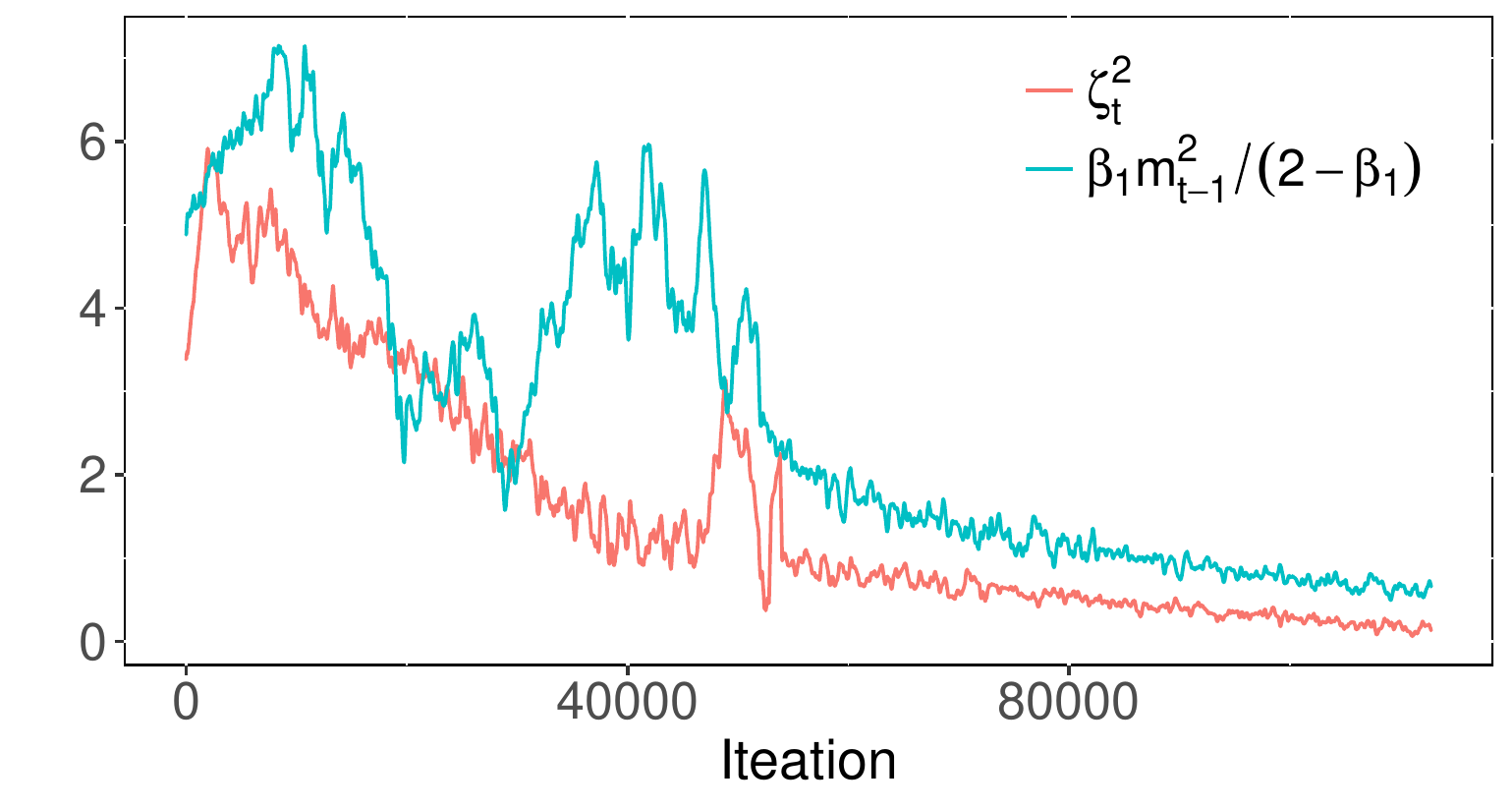}
    \caption{Empirical investigation of Assumption~\ref{asp:noise_decay}.}
    \label{fig: assumption}
\end{figure}

Assumption \ref{asp: gradient-unbiased-noise-ind} is a standard assumption in stochastic optimization~\citep{reddi2018convergence, savarese2021domain,guo2021stochastic}. As $\beta_1$ is always set as positive number close to $1$, Assumption~\ref{asp:noise_decay} basically requires that the gradient noise variance to be smaller than the second moment of $m$ when $t$ is very large. This is mild as 1) we can select mini-batch size to be large enough to satisfy it as the noise variance is inversely proportional to batch size~\citep{bubeck2014convex}. 2) The magnitudes of the variances of the stochastic gradients are usually much lower than that of the gradients~\citep{faghri2020study}. In Fig.~\ref{fig: assumption}, we report the values of $\zeta_t^2$ and $\beta_1m_{t-1}^2/(2-\beta_1)$ of AdaMomentum on the $5$-layer fully connected network with width $30$. From Fig.~\ref{fig: assumption}, one can observe that $\zeta_t^2$ is consistently lower than $\beta_1m_{t-1}^2/(2-\beta_1)$ as iteration becomes larger, which further validates Assumption~\ref{asp:noise_decay}. Then we can come to the following result.

\begin{proposition}\label{thm: generalize} Under Assumptions \ref{assu:escape1}-\ref{asp:noise_decay}, upon encountering a comparatively flat basin or asymmetric valley $\boldsymbol{\Omega}$, we have 
    \small
    \begin{equation*}
        \EE \left(\Gamma^{(\textsc{AdaMomentum})}\right) \ge \EE \left(\Gamma^{(\textsc{Adam})}\right).
    \end{equation*}
    \normalsize
\end{proposition}

When falling into a flat/asymmetric basin, AdaMomentum is more stable than Adam and will not easily escape from it. Combining the aforementioned results and the fact that minima at the flat or asymmetric basins tend to exhibit better generalization performance (as observed in~\citet{keskar2016large,he2019asymmetric,hochreiter1997flat,izmailov2018averaging,li2017visualizing}),  we are able to conclude that AdaMomentum is more likely to converge to minima that generalize better, which may buttress the  improvement of AdaMomentum in empirical performance. All the proofs in section~\ref{sec: escape_minima} are provided in Appendix~\ref{app: escape}.

%% file: convergence_convex.tex
\section{Convergence Analysis of AdaMomentum}\label{sec: convergence}

 In this section, we establish the convergence theory for AdaMomentum under both convex and non-convex object  function conditions. We omit the two bias correction steps in the Algorithm~\ref{alg: adamomentum} for simplicity and the following analysis can be easily adapted to the de-biased version as well.

\subsection{Convergence Analysis in Convex Optimization}
We analyze the convergence of AdaMomentum in convex setting utilizing the online learning framework~\citep{zinkevich2003online}. Given a sequence of convex cost functions $f_1(\theta), \cdots, f_T(\theta)$, the regret is defined as $R(T) = \sum_{t=1}^T [f_t(\theta_t) - f_t(\theta^*)]$, where $\theta^* = \argmin_{\theta}\sum_{t=1}^T f_t(\theta)$ is the optimal parameter and $f_t$ can be interpreted as the loss function at the $t$-th step. Then we have:
\begin{theorem}\label{thm: convex}
  Let $\{\theta_t\}$ and $\{v_t\}$ be the sequences yielded by AdaMomentum. Let $\alpha_t = \alpha/\sqrt{t}, \beta_{1,1} = \beta_1, 0<\beta_{1,t} \le \beta_1<1, v_t \le v_{t+1} $ for all $t \in [T]$ and $\gamma = \beta_1/{\sqrt{\beta_2}} < 1$. Assume that the distance between any $\theta_t$ generated by AdaMomentum is bounded, $\norm{\theta_m - \theta_n}_{\infty} \le D_{\infty}$ for any $m,n \in \{1, \cdots, T\}$. Then we have the following bound:
  \small
  \begin{align*}
    R(T) \le&  \frac{D_{\infty}^2\sqrt{T}}{2\alpha (1-\beta_1)}  \sum_{i=1}^d \sqrt{v_{T,i}}+ \frac{D_{\infty}^2}{2(1-\beta_1)} \sum_{t=1}^T \sum_{i=1}^d \frac{\beta_{1,t} \sqrt{v_{t,i}}}{\alpha_t} \\
    &+  \frac{\alpha \sqrt{1+\log T}}{(1-\beta_1)^3 (1-\gamma)\sqrt{1-\beta_2}} \sum_{i=1}^d \norm{g_{1:T, i}}_2.
  \end{align*}
  \normalsize
\end{theorem}
Theorem~\ref{thm: convex} implies that the regret of AdaMomentum can be bounded by $\tilde{O}\footnote{$\tilde{O}(\cdot)$ denotes $O(\cdot)$ with hidden logarithmic factors.}(\sqrt{T})$, especially when the data features are sparse as Section 1.3 in~\citet{duchi2011adaptive} and then we have $\sum_{i=1}^d\sqrt{v_{T,i}} \ll \sqrt{d}$ and $\sum_{i=1}^d \norm{g_{1:T,i}}_2 \ll \sqrt{dT}$. Imposing additional assumptions that $\beta_{1,t}$ decays exponentially and that the gradients of $f_t$ are bounded~\citep{DBLP:journals/corr/KingmaB14,liu2019variance}, we can obtain:
\begin{corollary}\label{cor: convex}
  Further Suppose $\beta_{1,t} = \beta_1 \lambda^t$ and the function $f_t$ has bounded gradients, $\norm{\nabla f_t(\theta)}_{\infty} \le G_{\infty}$ for all $\theta \in \RR^d$, AdaMomentum achieves the guarantee $R(T)/T = \tilde{O}(1/\sqrt{T})$ for all $T\ge 1$:
  \small
  \begin{align*}
    \frac{R(T)}{T} \le & \left[\frac{d\alpha \sqrt{1+\log T}}{(1-\beta_1)^3 (1-\gamma) \sqrt{(1-\beta_2)T}}+\frac{d D_{\infty}^2 }{2\alpha (1-\beta_1)\sqrt{T}}\right] \\
    &\cdot (G_{\infty}+\sqrt{{\epsilon}/{1-\beta_2}}) + \frac{d D_{\infty}^2 G_{\infty}\beta_1}{2\alpha (1-\beta_1)(1-\lambda)^2 T}.
  \end{align*}
  \normalsize
\end{corollary}
From Corollary~\ref{cor: convex}, the average regret of AdaMomentum converges to zero as $T$ goes to infinity. The proofs of Theorem~\ref{thm: convex} and Corollary~\ref{cor: convex}  are provided in Appendix~\ref{proof: convex}.

%% file: convergence_nonconvex_new.tex
\subsection{Convergence Analysis in Non-convex Optimization}
When  $f$ is non-convex and lower-bounded, we derive the non-asymptotic convergence rate of AdaMomentum.

\begin{theorem}\label{thm: nonconvex}
  Suppose that Assumptions~\ref{assu:escape2} and~\ref{asp: gradient-unbiased-noise-ind} hold. We denote $b_{u,t} = \sqrt{(1-\beta_2)/(\epsilon-\epsilon\beta_2^t)} \leq b_{u,1}, b_{l,t} = 1/\left[\sqrt{G^2(1-\beta^T)^2+\epsilon/(1-\beta_2)}(1-\beta_2^T)\right] \geq b_{l,T}$ where $\beta = \min_t \beta_{1,t}$. If there exists some $T_0 \lesssim 1/\alpha_T$, such that for all $t \geq T_0$ we have $\alpha_T \leq \alpha_t \leq (1-\beta_{1,t+1})\sqrt{b_{l,T}/(2L^2b_{u,1}^3)}$ and $\alpha_t \leq b_{l,t}/(2L b_{u,t}^2)$. With $\sum_{t=1}^{T} (1-\beta_{1,t})^2 = \eta(T)$, we have:
  \small
  \begin{equation*}
      \mathbb{E} \left[ \frac{1}{T+1} \sum_{t=0}^T \norm{\nabla f(\theta_t)}_2^2 \right] \leq \frac{1}{\alpha_T(T+1)}(Q_1 + Q_2 \eta(T))
  \end{equation*}
  \normalsize
  for some positive constants $Q_1, Q_2$ independent of 
  $d$ or $T$.
\end{theorem} 
 The conditions in Theorem~\ref{thm: nonconvex} are mild and reasonable, as in practice the momentum parameter for the first-order average $\beta_{1,t}$ is usually set as a large value, and meanwhile the step size $\alpha_t$ decays with time~\citep{chen2018on,guo2021stochastic,huang2021biadam}. In particular, we can use a setting with $(1-\beta_{1,t}) = 1/\sqrt{t}$ and $\alpha_t = \alpha/\sqrt{t}$ for some initial constant $\alpha$ to achieve the $O(\log(T)/\sqrt{T})$ convergence rate as in the following result.
\begin{corollary} \label{cor: nonconvex}
  When $1-\beta_{1,t}$ and $\alpha_t$ are further chosen to be in the scale of $O(1/{\sqrt{t}})$ with all assumptions in Theorem \ref{thm: nonconvex} hold, AdaMomentum satisfies:
  \small
  \begin{equation*}
      \mathbb{E} \left[ \frac{1}{T+1} \sum_{t=0}^T \norm{\nabla f(\theta_t)}_2^2 \right] \leq  \frac{1}{\sqrt{T}}(Q_1^* + Q_2^* \log(T)), 
  \end{equation*}
  \normalsize
  for some constants $Q_1^*, Q_2^*$ similarly defined in Theorem \ref{thm: nonconvex}.
\end{corollary}
Corollary~\ref{cor: nonconvex} manifests the $O(\log(T)/\sqrt{T})$ convergence rate of AdaMomentum under the nonconvex case. We refer readers to the detailed proof 
in Appendix~\ref{proof: nonconvex}.

%% file: experiment.tex
\begin{figure}[htbp]
  \centering
  \subfigure[$\alpha_{\text{Adam}}=0.1$.]{
    \includegraphics[width=.33\linewidth]{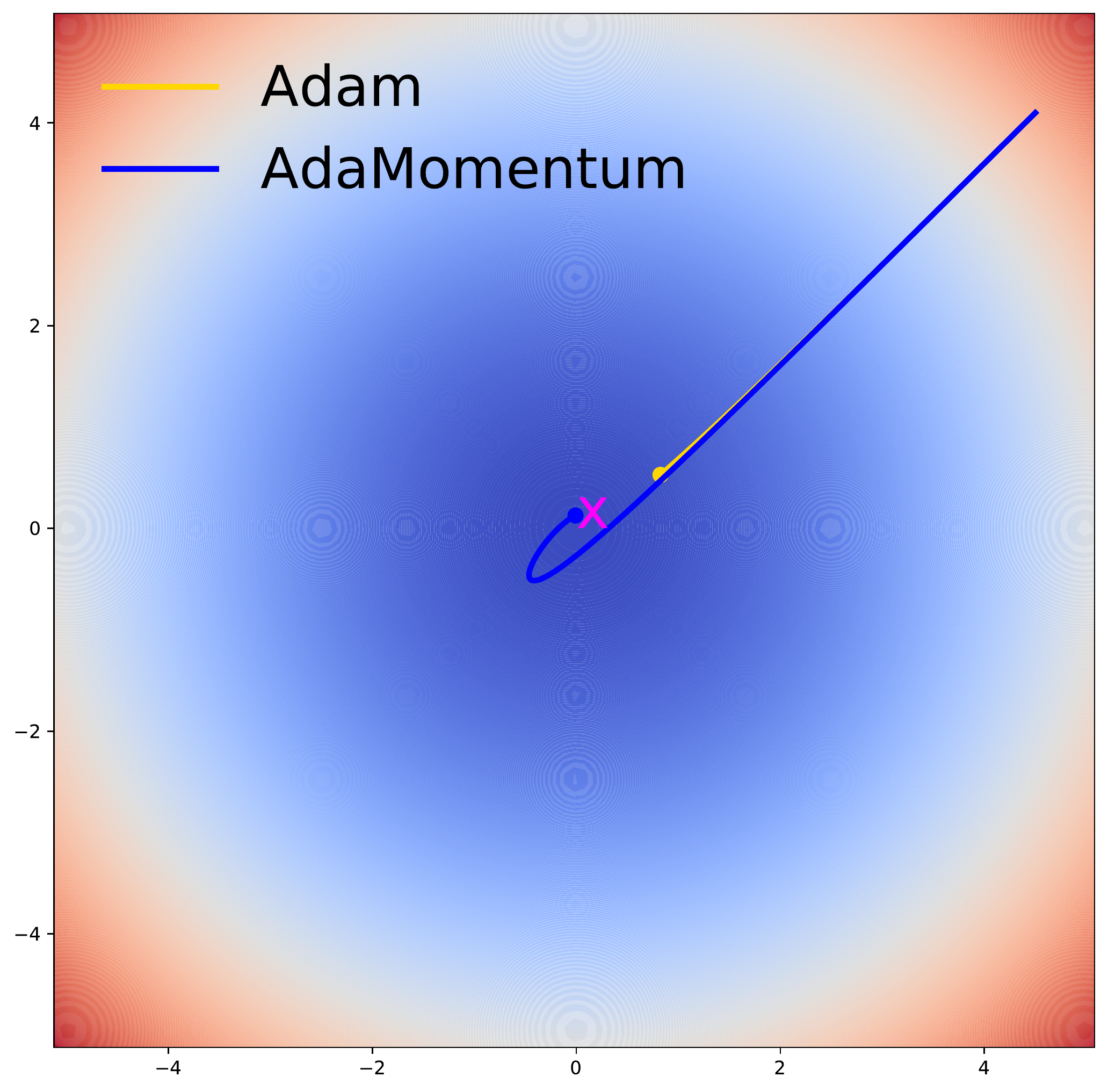}  
  }\hspace{-10mm}
  \hfill
  \subfigure[$\alpha_{\text{Adam}}=0.5$.]{
    \includegraphics[width=.33\linewidth]{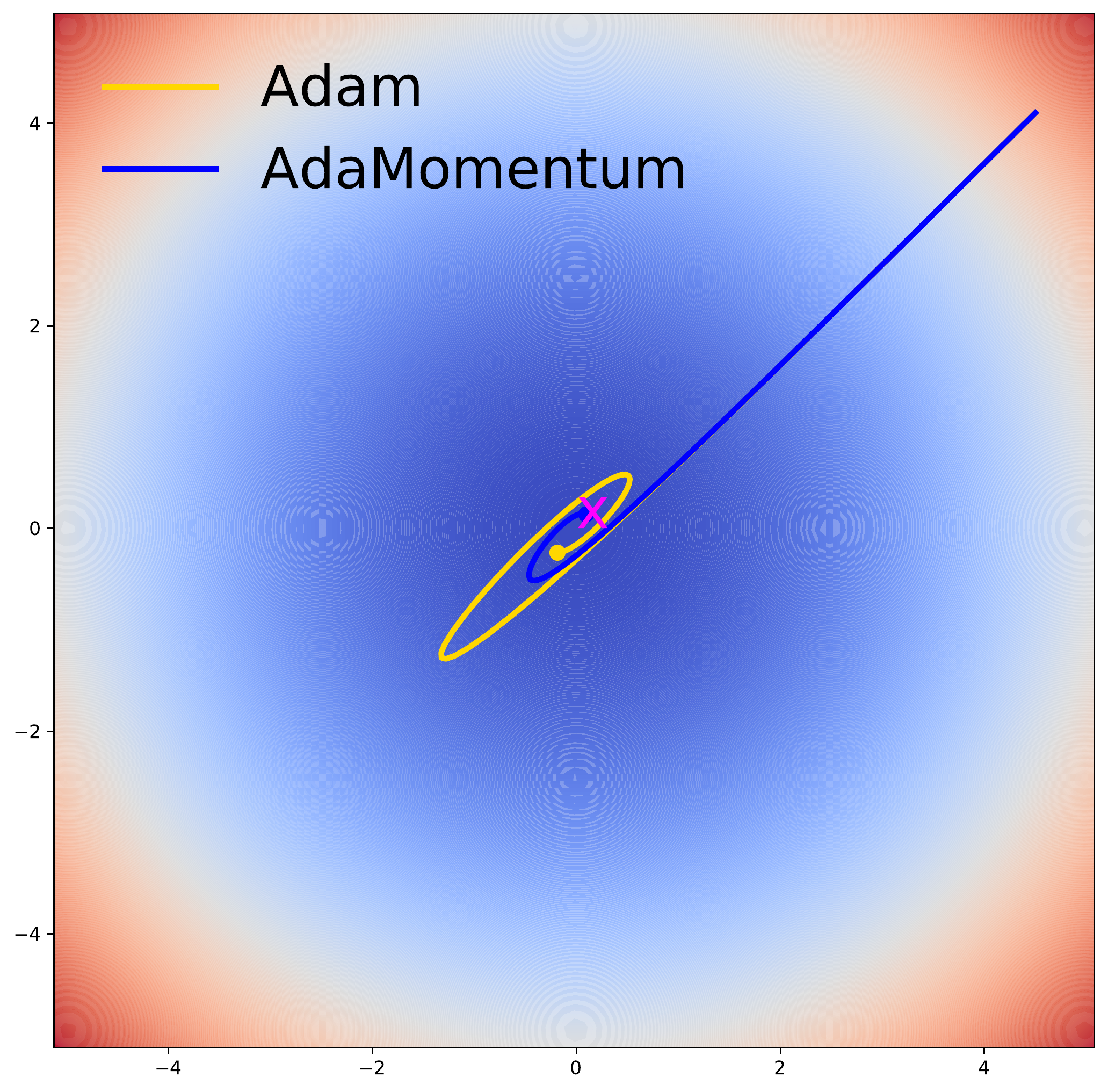}  
  }\hspace{-10mm}
  \hfill
  \subfigure[$\alpha_{\text{Adam}}=1.0$.]{
    \includegraphics[width=.33\linewidth]{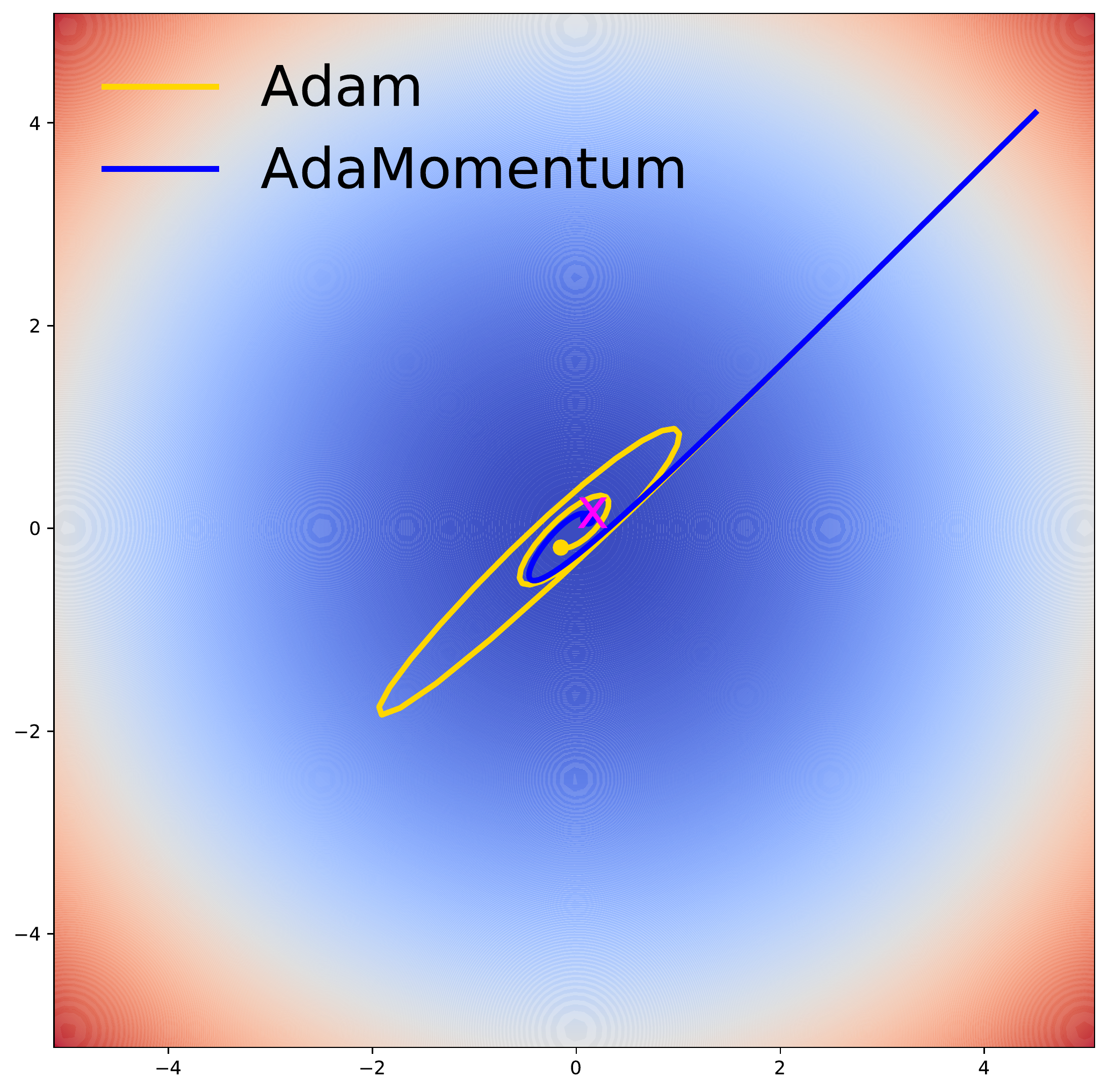}  
  }
  \caption{The optimization trajectories of Adamomentum and Adam on Sphere Function. The $\alpha$s of AdaMomentum are 0.1.}
  \label{fig: sphere}
\end{figure}
\begin{figure}[htbp]
  \centering
  \subfigure[Train Accuracy of VGGNet.]{
    \includegraphics[width=.47\linewidth]{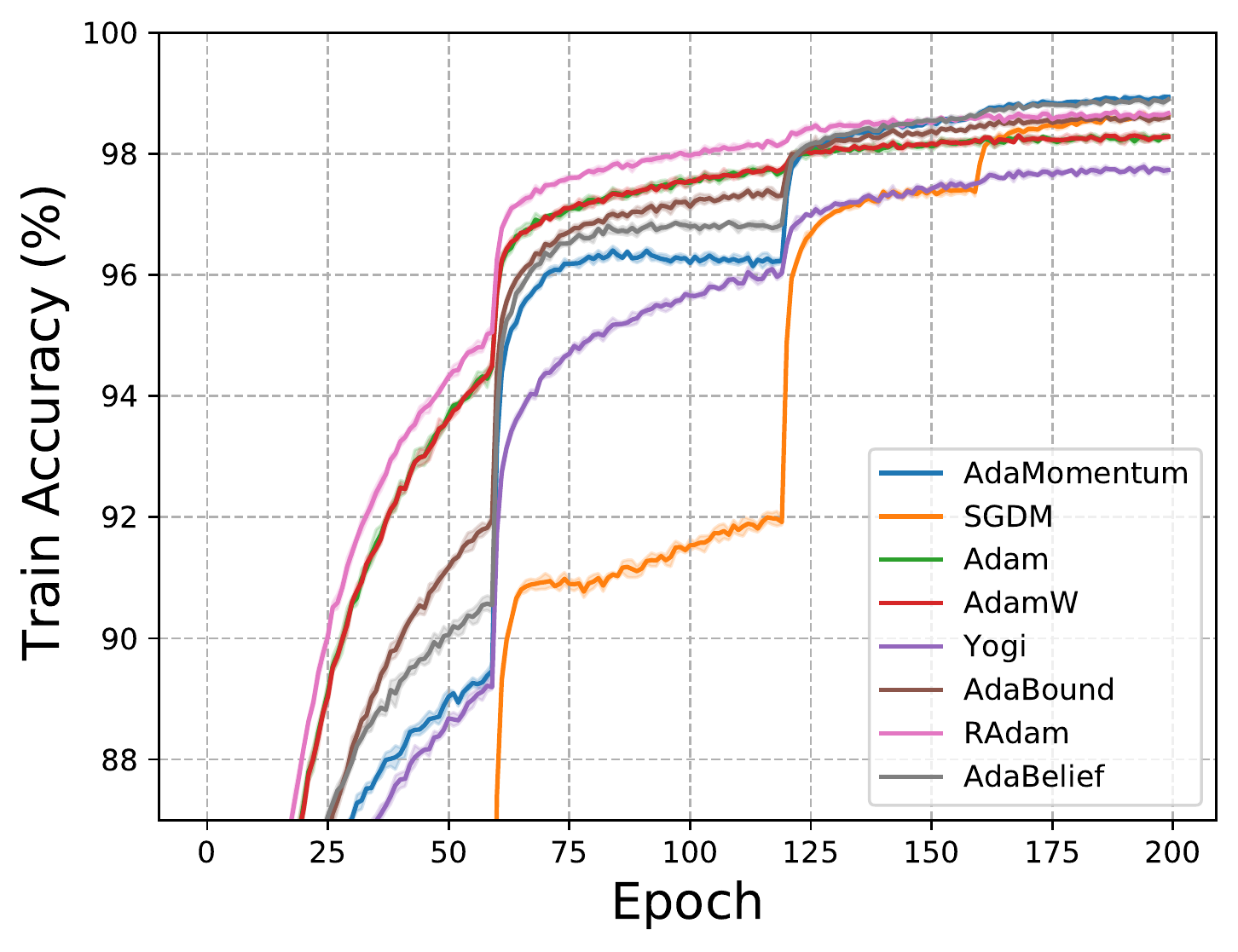}  
  }\hspace{-10mm}
  \hfill
  \subfigure[Test Accuracy of VGGNet.]{
    \includegraphics[width=.47\linewidth]{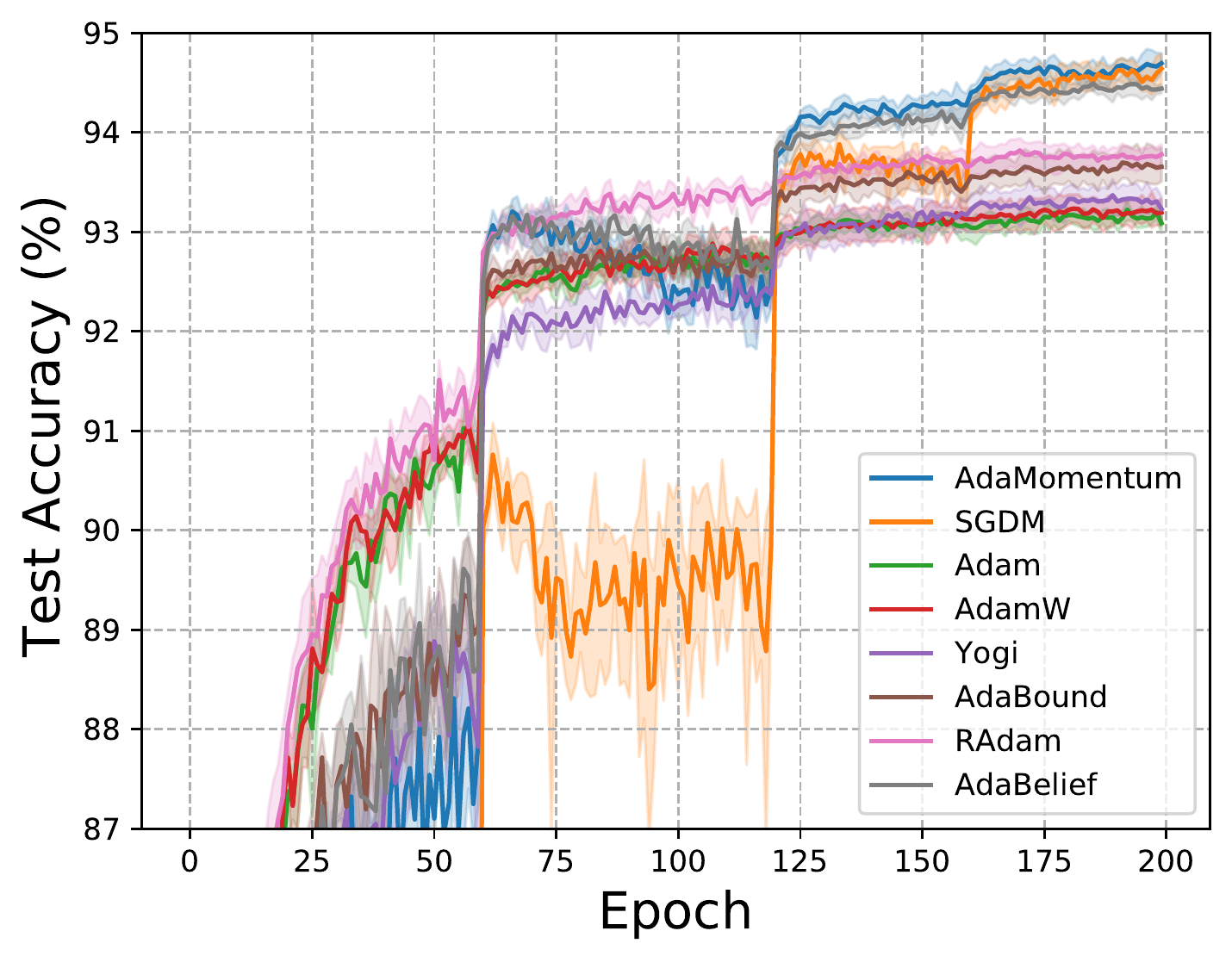}  
  }
  \quad
  \subfigure[Train Accuracy of ResNet.]{
    \includegraphics[width=.47\linewidth]{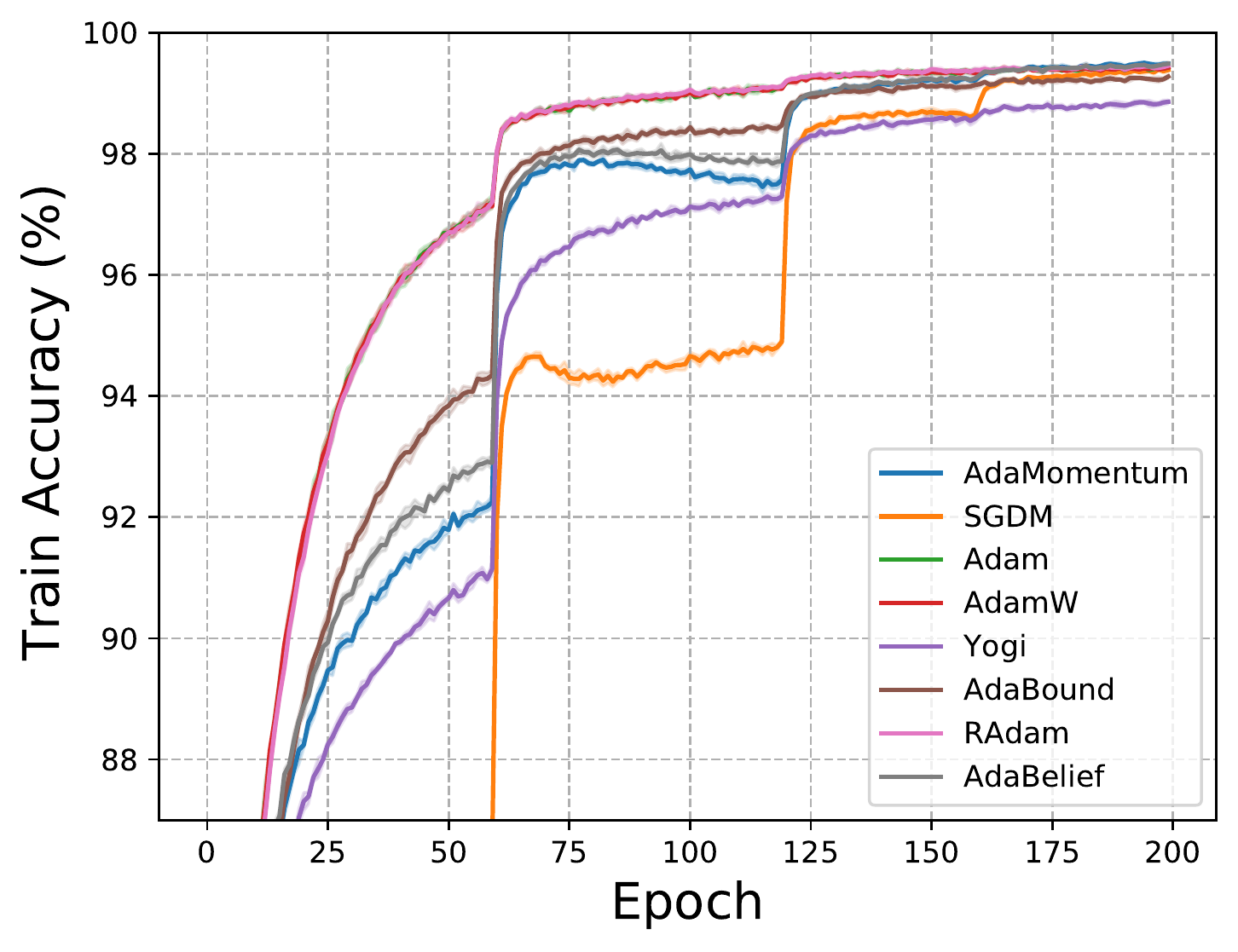}  
  }\hspace{-10mm}
  \hfill
  \subfigure[Test Accuracy of ResNet.]{
    \includegraphics[width=.47\linewidth]{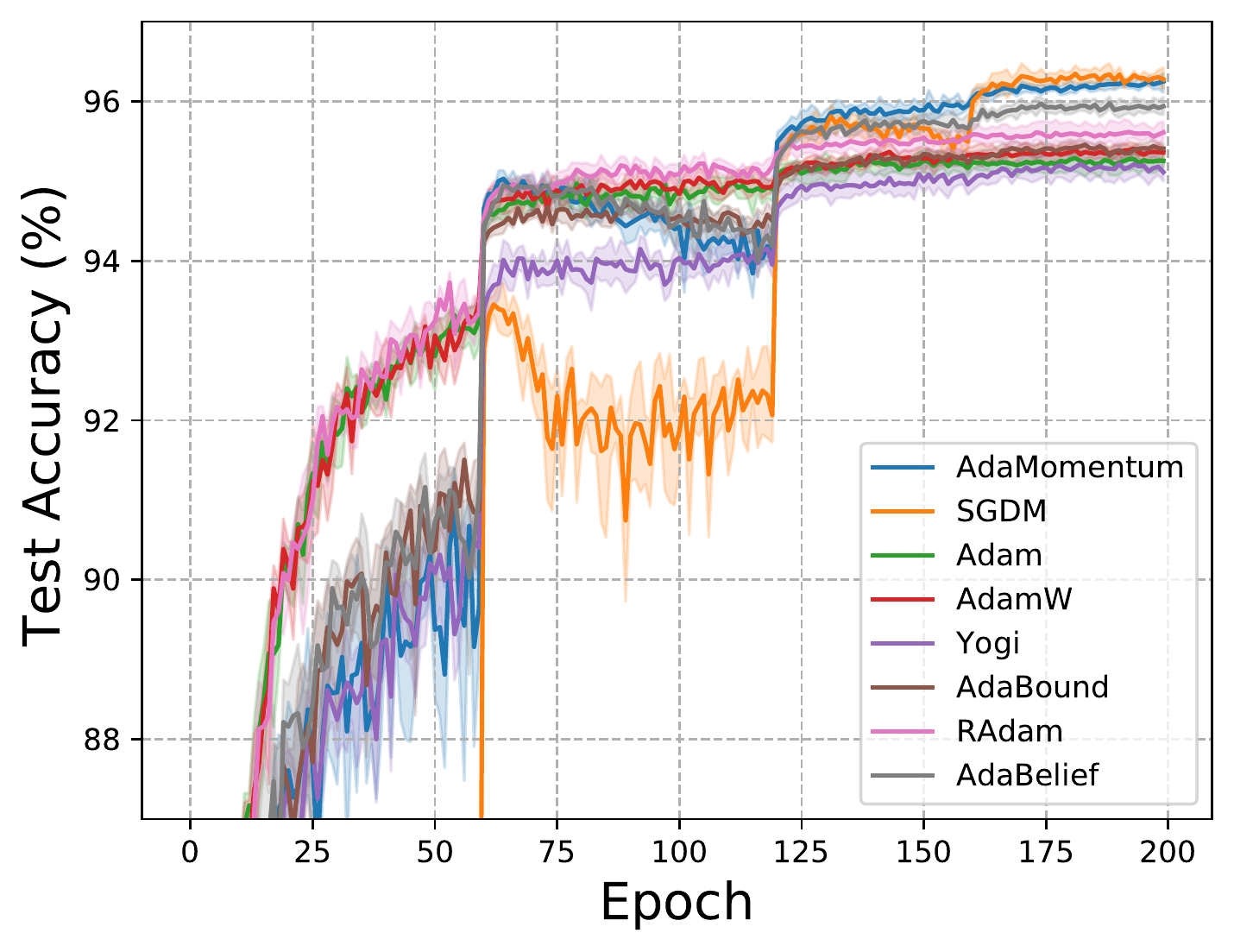}  
  }
  \quad
  \subfigure[Train Accuracy of DenseNet.]{
    \includegraphics[width=.47\linewidth]{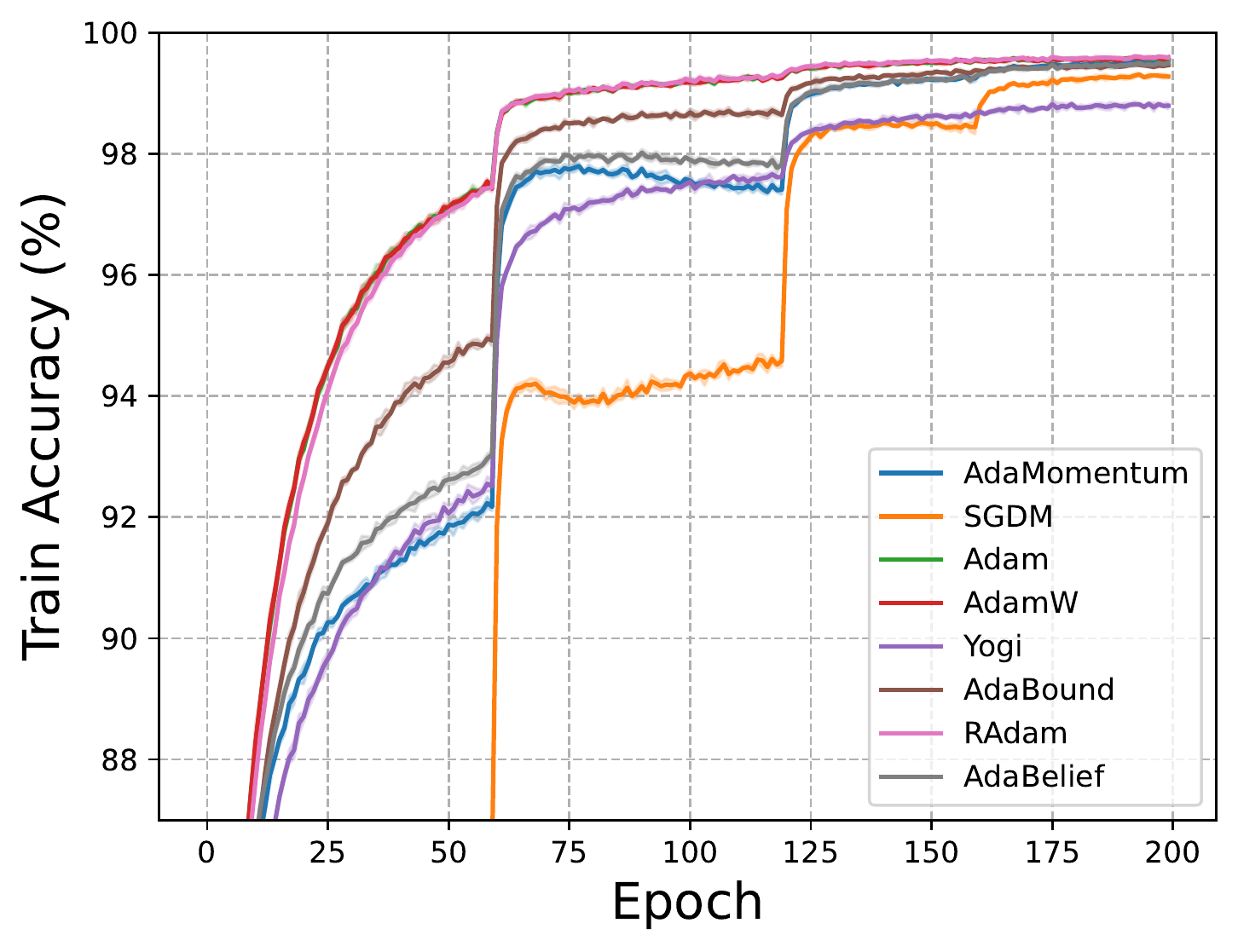}  
  }\hspace{-10mm}
  \hfill
  \subfigure[Test Accuracy of DenseNet.]{
    \includegraphics[width=.47\linewidth]{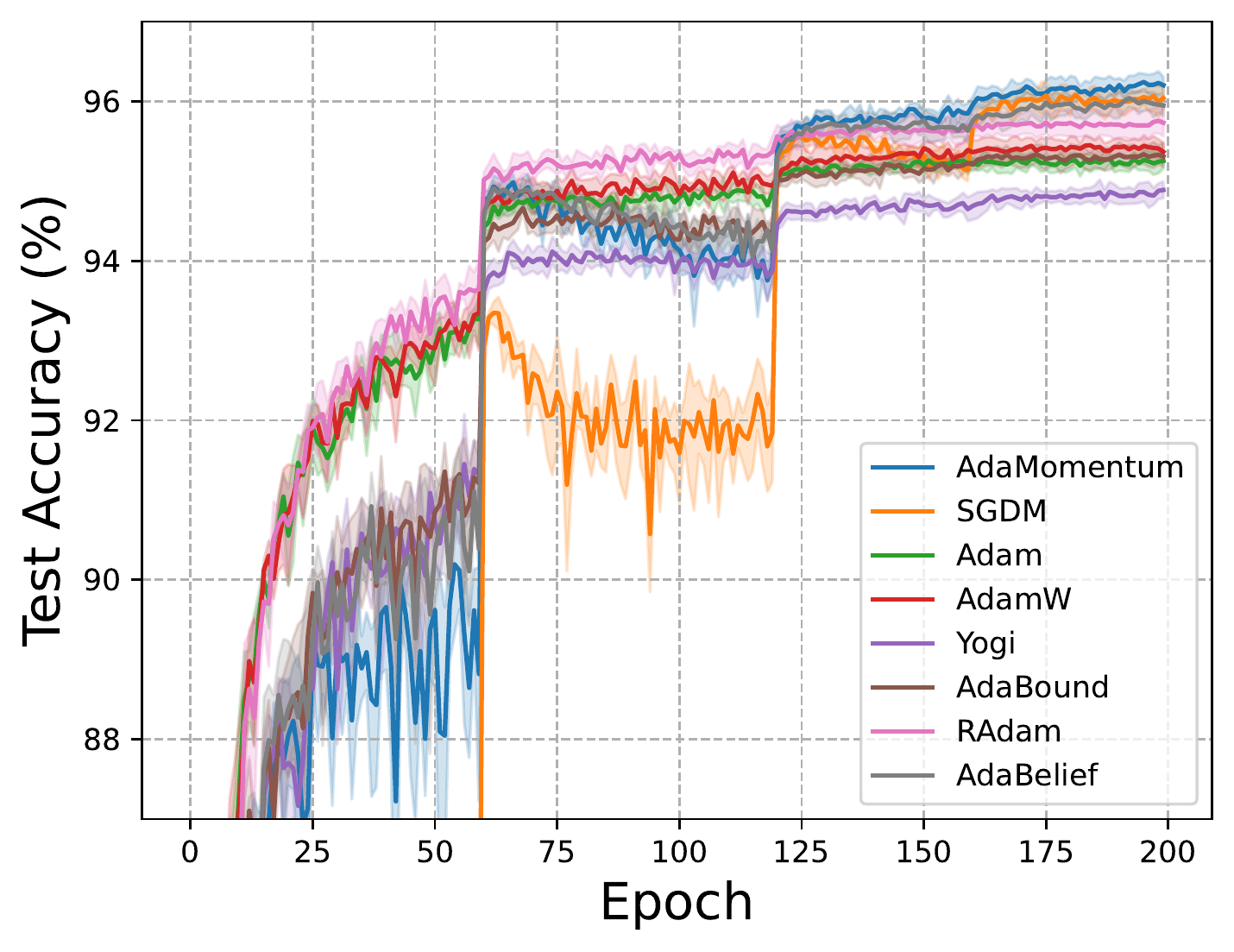}  
  }
  \caption{Train and test accuracy curves of all the compared optimizers on CIFAR-10~\cite{Krizhevsky09}.}
  \label{fig: cifar-10}
\end{figure}
\begin{table*}[htbp]
  \caption{Test accuracy ($\%$) of CNNs on CIFAR-10 dataset. The best in {\color{red} Red}  and second best in {\color{blue} blue}.}
  \label{table: cifar}
  \centering
  \scalebox{0.8}{
  \begin{tabular}{c|c|cccccccc}
  \toprule[1pt]
  \multirow{2}{*}{Architecture}& Non-adaptive& \multicolumn{7}{c}{Adaptive gradient methods} \\ 
  \cmidrule{2-9}
    & SGDM & Adam & AdamW & Yogi & AdaBound & RAdam & AdaBelief & {\bf Ours} \\ \midrule
   VGGNet-16& {\color{blue} 94.73$^{\small \pm 0.12 }$}& 93.29$^{\small \pm 0.10 }$& 93.33$^{\small \pm 0.15}$& 93.44$^{\small \pm 0.16}$& 93.79$^{\small \pm 0.17}$ & 93.90$^{\small \pm 0.10 }$ & 94.57$^{\small \pm 0.09 }$ & {\color{red} 94.80$^{\small \pm 0.10 }$}\\
   ResNet-34& {\color{red} 96.47$^{\small \pm 0.09 }$} & 95.39$^{\small \pm 0.11 }$ & 95.48$^{\small \pm 0.10 }$& 95.28$^{\small \pm 0.19 }$ & 95.51$^{\small \pm 0.07 }$ & 95.67$^{\small \pm 0.16 }$ & 96.04$^{\small \pm 0.07 }$& {\color{blue} 96.33$^{\small \pm 0.07}$} \\
   DenseNet-121& {\color{blue} 96.19$^{\small \pm 0.17 }$} & 95.35$^{\small \pm 0.09 }$ & 95.52$^{\small \pm 0.14 }$ & 94.98$^{\small \pm 0.13 }$ & 95.43$^{\small \pm 0.12 }$ & 95.82$^{\small \pm 0.19 }$ & 96.09$^{\small \pm  0.14}$ & {\color{red} 96.30$^{\small \pm 0.12}$} \\
   \bottomrule[1pt]
  \end{tabular}
  }
\end{table*}

\begin{figure}[htbp]
  \centering
  \subfigure[Adam.]{
    \includegraphics[width=.45\linewidth]{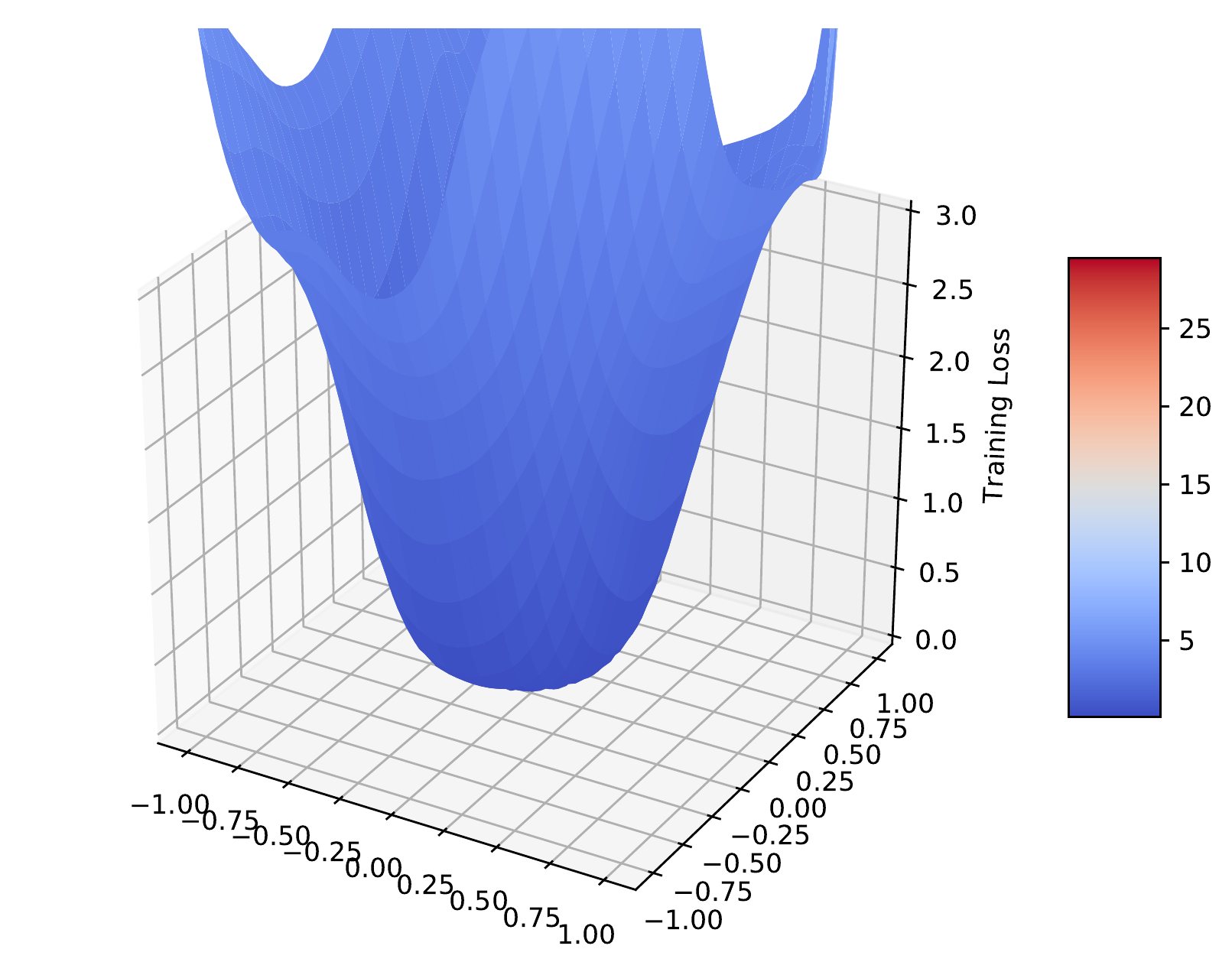}  
  }\hspace{-10mm}
  \hfill
  \subfigure[AdaMomentum.]{
    \includegraphics[width=.45\linewidth]{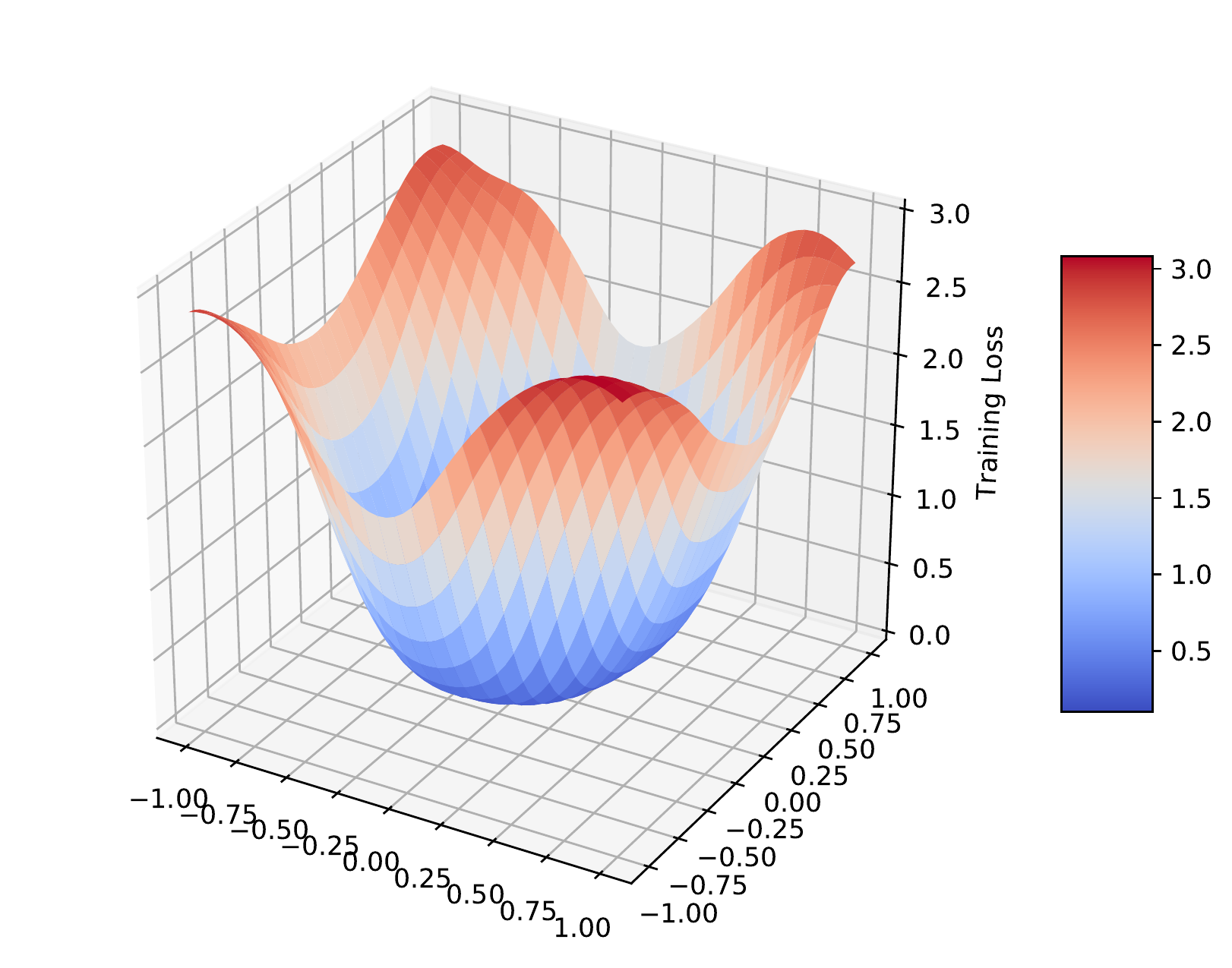}  
  }
  \caption{Comparison of the basins around the convergent points of ResNet-34 trained by Adam and AdaMomentum on CIFAR-10.}
  \label{fig: landscape}
\end{figure}
\section{Experiments}\label{sec: exp}

\subsection{2D Toy Experiment on Sphere Function}

We compare the optimization performance of AdaMomentum and Adam on 2D Sphere Function (bowl-shaped)~\cite{dixon1978global}: $f(x) = x_1^2 + x_2^2$. We omit the damping term $\epsilon$ in both two algorithms so the only difference is the $m_t$ and $g_t$ in the term $v_t$. We set the learning rate $\alpha$ of AdaMomentum as 0.1 and finetune that of Adam. We can observe from Figure~\ref{fig: sphere} that on one hand, when the $\alpha$ of Adam is the same as AdaMomentum, Adam is much slower than our Adamomentum in convergence; on the other hand, when we use larger $\alpha$ on Adam ($\alpha=0.5, 0.1$) it will oscillate much more violently. To summarize, the replacement of $g_t$ with $m_t$ in AdaMomentum makes the alteration of learning rate smoother and more suitable (see analysis in Sec~\ref{sec: curvation}) for the Sphere loss function. Despite the fact this is only a toy experiment, such local behavior of AdaMomentum and Adam may shed light to their performance difference in complex deep learning tasks in the sequel, as any complicated real function can be approximated using the compositions of sphere functions~\cite{yarotsky2017error}.

\subsection{Deep Learning Experiments}

We empirically investigate the performance of AdaMomentum in optimization, generalization and training stability. We conduct experiments on various modern network architectures for different tasks covering both vision and language processing area: {\bf 1)} image Classification on CIFAR-10~\cite{Krizhevsky09} and ImageNet~\cite{russakovsky2015imagenet} with CNN; {\bf 2)} language modeling on Penn Treebank~\cite{marcus-etal-1993-building} dataset using Long Short-Term Memory (LSTM)~\cite{hochreiter1997long}; {\bf 3)} neural machine translation on IWSTL'14 DE-EN~\cite{cettolo2014report} dataset employing Transformer; {\bf 4)} Generative Adversarial Networks (GAN) on CIFAR-10. We compare AdaMomentum with seven state-of-the-art optimizers: SGDM~\cite{sutskever2013importance}, Adam~\cite{DBLP:journals/corr/KingmaB14}, AdamW~\cite{loshchilov2017decoupled}, Yogi~\cite{reddi2018adaptive}, AdaBound~\cite{luo2019adaptive}, RAdam~\cite{liu2019variance} and AdaBelief~\cite{zhuang2020adabelief}. We perform a careful and extensive hyperparameter tuning (including learning rate, $\beta_2$, weight decay and $\epsilon$) for all the optimizers compared in each experiment and report their best performance. The detailed tuning schedule is summarized in Appendix~\ref{apd: details_exp} due to space limit. It is worth mentioning that in experiments we discover that setting $\alpha=0.001, \beta_1=0.9, \beta_2 = 0.999$ (the default setting for adaptive gradient methods in applied machine learing) works well in most cases. This elucidates that our optimizer is tuning-friendly, which reduces human labor and time cost and is crucial in practice. The mean results with standard deviations over $5$ random seeds are reported in all the following experiments except ImageNet. The source code of all the experiments are included in supplementary material.%

\begin{table}[htbp]
  \caption{Top-1 test accuracy (\%) on ImageNet.}
  \label{table: imagenet}
  \centering
  \scalebox{0.9}{
  \begin{tabular}{ccccccccc}
  \toprule[1pt]
   SGDM & Adam & Ours \\ \midrule
   70.73$\pm$0.07 &64.99$\pm$0.12 &{\bf 70.77$\pm$0.09} \\
   \bottomrule[1pt]
  \end{tabular}
  }
 
\end{table}
\begin{table*}[htbp]
  \caption{Test perplexity ($\downarrow$) results of LSTMs on Penn Treebank dataset.}
  \label{table: lstm}
  \centering
  \scalebox{0.85}{
  \begin{tabular}{c|ccccccccc}
  \toprule[1pt]
   Layer \#& SGDM & Adam & AdamW & Yogi & AdaBound & RAdam & AdaBelief & {\bf Ours} \\ \midrule
   1& 85.31$^{\small \pm 0.09}$&  84.55$^{\small \pm 0.10}$& 88.18$^{\small \pm 0.14}$& 86.87$^{\small \pm 0.14}$& 85.10$^{\small \pm 0.22}$ & 88.60$^{\small \pm 0.22}$ & 84.30$^{\small \pm 0.23}$ & {\bf 80.82$^{\bf\small \pm 0.19}$}\\
   2& 67.25$^{\small \pm 0.20}$ & 67.11$^{\small \pm 0.20}$ & 73.61$^{\small \pm 0.15}$& 71.54$^{\small \pm 0.14}$ & 67.69$^{\small \pm 0.24}$ & 73.80$^{\small \pm 0.25}$ & 66.66$^{\small \pm 0.11}$& {\bf 64.85$^{\bf\small \pm 0.09}$} \\
   3& 63.52$^{\small \pm 0.16}$ & 64.10$^{\small \pm 0.25}$ & 69.91$^{\small \pm 0.20}$ & 67.58$^{\small \pm 0.08}$ & 63.52$^{\small \pm 0.11}$ & 70.10$^{\small \pm 0.16}$ & 61.33$^{\small \pm 0.19}$ & {\bf 60.08$^{\bf\small \pm 0.11}$} \\
   \bottomrule[1pt]
  \end{tabular}
  }
  
  \end{table*}
\begin{table*}[htbp]
  \centering
    \caption{FID score ($\downarrow$) of GANs on CIFAR-10 dataset. $\dag$ is reported in \citet{zhuang2020adabelief}.}
    \label{table: gan}
    \scalebox{0.85}{
    \begin{tabular}{c|cccccccc}
    \toprule[1pt]
    Type of GAN& SGDM & Adam(W)  & Yogi & AdaBound & RAdam & AdaBelief & {\bf Ours}\\ \midrule
    DCGAN & 223.77$^{\small \pm 147.90}$  & 52.39$^{\small \pm 3.62}$  & 63.08$^{\small \pm 5.02}$  & 126.79$^{\small \pm 40.64}$ & 48.24$^{\small \pm 1.38}$  & 47.25$^{\small \pm 0.79}$  & {\bf 46.66$^{\small \pm 1.94}$} \\ 
     SNGAN &49.70$^{\small \pm 0.41}$$^\dag$ & 13.05$^{\small \pm 0.19}$$^\dag$ & 14.25$^{\small \pm 0.15}$$^\dag$ & 55.65$^{\small \pm 2.15}$$^\dag$& 12.70$^{\small \pm 0.12}$$^\dag$ & 12.52$^{\small \pm 0.16}$$^\dag$ & {\bf 12.06$^{\small \pm 0.21}$} \\
     BigGAN & 16.12$^{\small \pm 0.33}$  & 7.24$^{\small \pm 0.08}$ &  7.38$^{\small \pm 0.04}$  &  14.81$^{ \small \pm 0.31}$ &  7.17$^{\pm 0.06}$  &  7.22$^{\pm 0.09}$  & {\bf 7.16$^{\pm 0.05}$}\\
     \bottomrule[1pt]
    \end{tabular}
    }
\end{table*}

\subsubsection{CNN for Image Classification}

\paragraph{CIFAR-10}

We experimented with three prevailing deep CNN architectures: VGG-16~\cite{simonyan2014very}, ResNet-34~\cite{he2016deep} and DenseNet-121~\cite{huang2017densely}. In each experiment we train the model for $200$ epochs with batch size $128$ and decay the learning rate by $0.2$ at the $60$-th, $120$-th and $160$-th epoch. We employ label smoothing technique~\cite{szegedy2016rethinking} and the smoothing factor is choosen as $0.1$.  Figure~\ref{fig: cifar-10} displays the training and testing results of all the compared optimizers . As indicated, both the training accuracy and the testing accuracy using AdaMomentum can be improved as fast as with other adaptive gradient methods, being much faster than SGDM, especially before the third learning rate annealing. In testing phase, AdaMomentum can exhibit performance as good as SGDM and far exceeds other baseline adaptive gradient methods, including the recently proposed AdaBelief~\cite{zhuang2020adabelief} optimizer. This contradicts the result reported in~\citet{zhuang2020adabelief}, where they claim AdaBelief can be better than SGDM. This largely stems from the fact that \citet{zhuang2020adabelief} did not take an appropriate stepsize annealing strategy or tune the hyperparameters well. Training $200$ epochs with ResNet-34 on CIFAR-10, our experiments show that AdaMomentum and SGDM can reach over $96\%$ accuracy, while in~\citet{zhuang2020adabelief} the accuracy of SGDM is only around $94\%$. 

We further visualize the basins of the convergent minima of the models trained by Adam and AdaMomentum respectively in Figure~\ref{fig: landscape}. We trained ResNet-34 on CIFAR-10 using radndom seed $0$ for $200$ epochs, and depict the 3D loss landscapes along with two random directions~\cite{li2017visualizing}. The landscape of Adam is cropped along the z axis to $3$ for comparison. Obviously seen from Figure~\ref{fig: landscape}, the basin of Adamomentum is much more flat than that of Adam, which verifies our theoretical argument in Sec~\ref{sec: escape_minima} that AdaMomentum is more likely to converge to flat minima.

\paragraph{ImageNet}

To further corroborate the effectiveness of our algorithm on more comprehensive dataset, we perform experiments on ImageNet ILSVRC 2012 dataset~~\cite{russakovsky2015imagenet} utilizing ResNet-18 as backbone network. We execute each optimizer for $100$ epochs utilizing cosine annealing strategy, which can exhibit better performance results than step-based decay strategy on ImageNet~\cite{loshchilov2016sgdr,ma2020apollo}. Each optimizer runs three times independently. As indicated in Table~\ref{table: imagenet}, AdaMomentum far exceeds Adam in Top-1 test accuracy and even performs slightly better than SGDM. 

\subsubsection{LSTM for Language Modeling}

We implement LSTMs with $1$ to $3$ layers on Penn Treebank dataset, where adaptive gradient methods are the main-stream choices (much better than SGD). In each experiment we train the model for $200$ epochs with batch size of $20$ and decay the learning rate by $0.1$ at $100$-th and $145$-th epoch. Test perplexity (the lower the better) is summarized in Table~\ref{table: lstm}. Clealy observed from Table~\ref{table: lstm}, AdaMomentum achieves the lowest perplexity in all the settings and consistently outperform other competitors by a considerable margin. The training and testing perplexity curve is given in Figure~\ref{fig: lstm-train} and~\ref{fig: lstm-test} in Appendix~\ref{apd: details_exp} due to space limit. Particularly on $2$-layer and $3$-layer LSTM, AdaMomentum maintains both the fastest convergence and the best performance, which substantiates its superiority.

\subsubsection{Transformer for Neural Machine Translation}

\setlength\tabcolsep{3pt}
\begin{table}[htbp]
    \caption{BLEU score ($\uparrow$) on IWSTL'14 DE-EN dataset.}
    \label{table: transformer}
    \centering
    \scalebox{0.85}{
    \begin{tabular}{ccccc}
    \toprule[1pt]
      SGDM & Adam& AdamW  & AdaBelief & {\bf Ours} \\ \midrule
      28.22$\pm$0.21 & 30.14$\pm$1.39 &35.62$\pm$0.11  & 35.60$\pm$0.11 & {\bf 35.66$\pm$0.10}\\
    \bottomrule[1pt]
    \end{tabular}
    }
\end{table}
Transformers have been the dominating architecture in NLP and adaptive gradient methods are usually adopted for training owing to their stronger ability to handle attention-models~\cite{zhang2019adaptive}. To test the performance of AdaMomentum on transformer, we experiment on IWSTL'14 German-to-English with the Transformer \textit{small} model adapting the code from fairseq package.\footnote{\url{https://github.com/pytorch/fairseq}} We set the length penalty as $1.0$, the beam size as $5$, warmup initial stepsize as $10^{-7}$ and the warmup updates iteration number to be $8000$. We train the models for $55$ epochs and the results are reported according to the average of the last $5$ checkpoints. As shown in Table~\ref{table: transformer}, our optimizer achieves the highest average BLEU score with the lowest variance.

\subsubsection{Generative Adversarial Network}
Training of GANs is extremely unstable. To further study the optimization ability and numerical stability of AdaMomentum, we experiment on three types of GANs: Deep Convolutional GAN (DCGAN)~\cite{radford2015unsupervised}, Spectral normalized GAN (SNGAN)~\cite{miyato2018spectral} and BigGAN~\cite{brock2018large}. For the generator and the discriminator network, we adopt CNN for DCGAN and ResNets for SNGAN and BigGAN. The BigGAN training is assited with consistency regularization~\cite{Zhang2020Consistency} for better performance. We train DCGAN for $200000$ iterations and the other two for $100000$ iterations on CIFAR-10 with batch size $64$. The learning rates for the generator and the discriminator network are both set as $0.0002$. For AdaMomentum all the other hyperparameters are set as default values. Experiments are run $5$ times independently and we report the mean and standard deviation of Frechet Inception Distance (FID, the lower the better)~\citet{heusel2017gans} in Table~\ref{table: gan}. From Table~\ref{table: gan} it is reasonable to draw the conclusion that AdaMomentum outperforms all the best tuned baseline optimizers for all the GANs by a considerable margin, which validates its outstanding optimization ability and numerical stabiliy. Here Adam equals AdamW because the optimal weight decay parameter value is $0$.

%% file: conclusion.tex
\section{Conclusion}
In this work, we rethink the formulation of Adam and proposed AdaMomentum as a new optimizer for machine learning adopting a twofold EMA approach. We illustrate that AdaMomentum is more fit to general loss curve than Adam and theoretically demonstrate why AdaMomentum outperforms Adam in generalization. We further validates the superiority of AdaMomentum through extensive and a broad range of experiments. Our algorithm is simple and effective with four key advantages: {\bf 1)} maintaining fast convergence rate; {\bf 2)} closing the generalization gap between adaptive gradient methods and SGD; {\bf 3)} applicable to various tasks and models; {\bf 4)} introducing no additional parameters and easy to tune. Combination of AdaMomentum with other techniques such as  Nesterov's accelerated gradient~\citep{dozat2016incorporating} may be of independent interest in the future.

%% file: appendix_escapeminima.tex
\section{Technical details of Subsection~\ref{sec: escape_minima}}\label{app: escape}
Here we provide more construction details and technical proofs for the L\'evy-driven SDE in Adam-alike adaptive gradient algorithm \eqref{adam_alike_eq}. In the beginning we introduce a detailed derivation of the process \eqref{sde_eq} as well as its corresponding escaping set $\Upsilon$ in definition \ref{def: escaping}. Then we give some auxiliary theorems and lemmas, and summarize the proof of Lemma \ref{thm: escape_time}. Finally we prove the proposition \ref{thm: escape_time} and give a more detailed analysis of the conclusion that the expected escaping time of AdaMomentum is longer than that of Adam in a comparatively flat basin.
\subsection{Derivation of the L\'evy-driven SDE \eqref{sde_eq}}
To derive the SDE of Adam-alike algorithms \eqref{adam_alike_eq}, we firstly define $m_t^\prime = \beta_1 m_{t-1}^\prime + (1 - \beta_1) \nabla f(\theta_t)$ with $m_0^\prime = 0$. Then by the definition it holds that
\begin{align*}
    m_t^\prime - m_t = (\beta_1 - 1) \sum_{i=0}^t \beta_1^{t-i} \zeta_t.
\end{align*}
Following \citet{simsekli2019tail}, the gradient noise $\zeta_t$ has heavy tails in reality and hence we assume that $\frac{1}{1-\beta_1}(m_t^\prime - m_t)$ obeys $\mathcal{S}\tilde{\alpha}\mathcal{S}$ distribution with time-dependent covariance matrix $\Sigma_t$. Since we can formulate \eqref{adam_alike_eq} as
\begin{align}
   \theta_{t+1} = \theta_{t} - \alpha \frac{m_t^\prime}{z_t} + \alpha\frac{(m_t^\prime - m_t)}{z_t} \text{  where  } z_t = (1-\beta_1^t)\sqrt{\frac{v_t}{(1-\beta_2^t)}}, \label{eq: discrete_sde}
\end{align}
and we can replace the term $(m_t^\prime - m_t)$ by $\alpha^{-\frac{1}{\tilde\alpha}} (1 - \beta_1^t) \Sigma_t S$ where each coordinate of $S$ is independent and identically distributed as $\mathcal{S}\tilde{\alpha}\mathcal{S}(1)$ based on the property of centered symmetric $\tilde\alpha$-stable distribution. Let $R_t = \text{diag}(\sqrt{\frac{v_t}{(1-\beta_2^t)}})$, and we further assume that the step size $\alpha$ is small, then the continuous-time version of the process \eqref{eq: discrete_sde} becomes the following SDE:
\begin{align}
    d\theta_t = -R_t^{-1} \frac{m_t^\prime dt }{ (1-\beta_1^t)}+ \alpha^{1-\frac{1}{\tilde\alpha}} R_t^{-1} \Sigma_t d L_t, \; dm_t = \beta_1(\nabla f(\theta_t) - m_t), \; dv_t = \beta_2(k_t^2 - v_t). \nonumber
\end{align}
After replacing $m_t^\prime$ with $m_t$ for brevity, we get the SDE \eqref{sde_eq} consequently.
\subsection{Proof of Lemma \ref{thm: escape_time}}
To prove Lemma~\ref{thm: escape_time}, we first introduce Theorem~\ref{thm: escape_pf}.
\begin{theorem}
\label{thm: escape_pf}
	Suppose Assumptions \ref{assu:escape1}-\ref{assu:escape3} hold. We define $\kappa_1 = \frac{c_1 L}{v_-\lvert \tau_m - 1\rvert}$ and $\kappa_2 = \frac{2\mu\tau}{\beta_1 v_+ + \mu\tau}\left(\beta_1-\frac{\beta_2}{4}\right)$ with a constant $c_1$. Let $\upsilon^{\tilde \alpha + 1} = \Theta(\tilde \alpha)$, $\rho_0=\frac{1}{16(1+c_2)}$
	and $\ln{\left(\frac{2\Delta}{\mu \upsilon^{1/3}}\right)} \leq \kappa_2 \upsilon^{-1/3}$ where $\Delta = f(\theta_0) - f(\theta^*)$ and a constant $c_2$.
	Then for any  $\theta_{0}\!\in \boldsymbol{\Omega}^{-2\upsilon^\gamma}$, $u>\!-1$, $\upsilon\in(0,\upsilon_0]$, $\gamma\in (0,\gamma_0]$ and $\rho\in(0,\rho_0] $ satisfying $\upsilon^{\gamma}\leq \rho_0$ and $\lim_{\upsilon\rightarrow 0}\rho = 0$, the Adam-alike algorithm in \eqref{adam_alike_eq} obey 
	\begin{equation*} 
	{ \frac{1-\rho }{1+u+\rho} }\leq	\mathbb{E}\left[{\exp{\left(-um(\Upsilon) \Theta(\upsilon^{-1}) \Gamma \right)}}\right] \leq  {\frac{1+\rho }{1+u-\rho}}.
	\end{equation*}
\end{theorem}
From Theorem \ref{thm: escape_pf}, by setting $\upsilon$ small, it holds that for any adaptive gradient algorithm the upper and lower bounds of its expected escaping time $\Gamma$ is at the order of $\left(\frac{\upsilon}{m(\Upsilon)}\right)$, which directly implies Lemma \ref{thm: escape_time} conclusively. Therefore, it suffices to validate Theorem \ref{thm: escape_pf}. 

The proof of Theorem~\ref{thm: escape_pf} is given in Section~\ref{sec: pf_for_escape}. Before we proceeed, we first provide some prerequisite notations in Section~\ref{sec: pre-notations} and list some useful theorems and lemmas in Section~\ref{sec: aux_for_escape}.

\subsubsection{Preliminaries}\label{sec: pre-notations}
For analyzing the uniform L\'evy-driven SDEs in \eqref{sde_eq}, we first introduce the L\'evy process $L_t$ into two components $\xi_t$ and $\varepsilon_i$, namely
\begin{equation}\label{decompositionlevy_eq}
L_{t} = \xi_{t} + \varepsilon_{t},
\end{equation}
whose characteristic functions are respectively defined as

\begin{align*}
\mathbb{E} \left[e^{i\langle \lambda, \xi_{t}\rangle }\right] =&e^{t  {\int}_{\mathbb{R}^{d}\setminus\{\bm{0}\}} \varepsilon I\left\{\norm{ y}_2 \leq \frac{1}{\upsilon^\delta}\right\}\nu(dy)}, \\
\mathbb{E} \left[e^{i\langle \lambda, \varepsilon_{t}\rangle }\right] =&e^{t{\int}_{\mathbb{R}^{d}\setminus\{\bm{0}\}} \varepsilon I\left\{\norm{ y}_2 \leq \frac{1}{\upsilon^\delta}\right\} \nu(dy)},
\end{align*}

where $\varepsilon = e^{i \langle \lambda, y\rangle} - 1 - i \langle \lambda, y\rangle I\left\{\norm{ y}_2 \leq 1\right\}$ with $\upsilon$ defined in \eqref{sde_eq} and a constant $\delta$ s.t. $\upsilon^{-\delta} < 1$. Accordingly, the L\'evy measure $\nu$ of the stochastic processes $\xi$ and $\varepsilon$ are 
\begin{align*}
    \nu_\xi = \nu\left( A \cap \left\{\norm{ y}_2 \leq \frac{1}{\upsilon^\delta}\right\}\right), \quad \nu_\varepsilon = \nu\left( A \cap \left\{\norm{y}_2 \geq \frac{1}{\upsilon^\delta}\right\}\right), \ \text{ where  } A \in \mathcal{B}(\mathbb{R}^d).
\end{align*}
Besides, for analysis, we should consider affects of the L\'evy motion $L_t$ to the L\'evy-driven SDE of Adam variants. Here we define the L\'evy-free SDE accordingly:
\begin{equation}\label{deterministicversion_adam_eq}
\begin{cases}
d \hat\theta_{t} = & -\mu_{t} \hat{Q}_{t}^{-1}\hat{m}_t, \\
d \hat{m}_t = &  \beta_1(\nabla f(\hat\theta_{t})  - \hat{m}_t),\\
d \hat{v}_t = & \beta_2( \nabla (f\hat{\theta}_{t})^2 -\hat{v}_t).\\
\end{cases}
\end{equation}
where $\hat{Q}_{t} = \text{diag}(\sqrt{\hat{v}_t})$.
\subsubsection{Auxiliary theorems and lemmas} 
\label{sec: aux_for_escape}

\begin{theorem}[\citet{zhou2020generalizationdeep}]
\label{thm: linearconvergenceadam}
	Suppose Assumptions \ref{assu:escape1}-\ref{assu:escape3} hold. Assume the sequence $\{(\hat\theta_{t},\hat{m}_{t},\hat{v}_{t})\}$ are produced by ~\eqref{deterministicversion_adam_eq}. Let $\hat{s}_{t}=\frac{h_t}{q_{t}}\left(\sqrt{\omega_{t}\hat{v}_{t}} \right)$ with $h_t= \beta_1$, $q_{t}=(1- \beta_1^t)^{-1}$ and $\omega_{t}=(1- \beta_2^t)^{-1}$.  We define $\norm{x}_{y}^2=\sum_i y_i x_i^2$.  Then for  L\'evy-driven Adam SDEs in \eqref{deterministicversion_adam_eq}, its Lyapunov function 
	$\mathcal{L}(t) = f(\hat\theta_{t}) -f(\hat\theta^*)+ \frac{1}{2} \norm{\hat{m}_t}_{\hat{s}_t^{-1}}$ with the optimum solution $\theta^*$ in the current local basin $\boldsymbol{\Omega}$ obeys 
\begin{align*}
	\mathcal{L}(t)\leq \Delta \exp\left(-\frac{2\mu\tau}{\beta_1 v_+ +\mu\tau}   \left(\beta_1 -\frac{\beta_2}{4} \right)t \right),
\end{align*}
	where $\Delta=f(\hat\theta_{0}) -f(\hat\theta^*)$ due to $\hat{m}_0 = 0$. The sequence $\{\hat\theta_t\}$ produced by \eqref{deterministicversion_adam_eq} obeys
\begin{align*}
	\norm{ \hat\theta_t - \theta^* }_2^2  \leq \frac{2\Delta}{\mu} \exp\left(-\frac{2\mu\tau}{ \beta_1 v_+  +\mu\tau}   \left(\beta_1 -\frac{\beta_2}{4} \right)t   \right).
\end{align*}
\end{theorem}
\begin{lemma}[\citet{zhou2020generalizationdeep}]
\label{lemma2}
(1) The process $\xi$ in the L\'evy process decomposition can be decomposed into two processes $\hat\xi$ and linear drift, namely,
\begin{equation}\label{adasfcsa}
    \xi_{t}=\hat\xi_{t}+ \mu_\upsilon t,
\end{equation}
where $\hat\xi$ is a zero mean L\'evymartingale with bounded jumps. \\
(2) Let $\delta \in (0,1), \mu_\upsilon = \mathbb{E}(\xi_1)$ and $T_\upsilon = \upsilon^{-\theta}$ for some $\theta > 0, \rho_0 = \rho_0(\delta) = \frac{1-\delta}{4} > 0$ and $\theta_0 = \theta_0(\delta) = \frac{1-\delta}{3} > 0$. Suppose $\upsilon$ is sufficiently small such that $\Theta(1) \leq \upsilon^{-\frac{1-\delta}{6}}$ and $\upsilon^{-\rho} - 2(C + \Theta(1)) \upsilon^{\frac{7}{6}(1-\delta)+\frac{\rho}{2}} \geq 1$ with a constant $C = \lvert \int_{0<u \leq 1} u^2 d \Theta(u) \rvert \in (0,+\infty)$. Then for all $\delta \in (0,\delta_0), \theta \in (0, \theta_0)$ there are $p_0 = p_0(\delta) = \frac{\delta}{2}$ and $\upsilon_0 = \upsilon_0(\delta,\rho)$ such that the estimates
\begin{align*}
    \norm{\upsilon \xi_{T_\upsilon}}_2 = \upsilon \norm{ \mu_\upsilon }_2T_\upsilon < \upsilon^{2 \rho}
    \text{   and   } P([\upsilon \xi]_{T_\upsilon}^d \geq \upsilon^\rho) \leq \exp(-\upsilon^{-p}),
\end{align*}
hold for all $p \in (0, p_0]$ and $\upsilon \in (0, \upsilon_0]$
\end{lemma}

\begin{lemma}[\citet{zhou2020generalizationdeep}]
\label{lemma3}
Let $\delta\in (0,1)$ and $g^{t}_{t\geq 0}$ be a bounded adapted c\.{a}dl\.{a}g stochastic process with values in  $\mathbb{R}^{d}$, $T_\upsilon=\upsilon^{-\theta}$, $\theta>0$. Suppose  $\sup_{t \geq 0} \|g^t\|$ is well bounded. Assume $\rho_0=\rho_0(\delta)=\frac{1-\delta}{16}>0$, $\theta_0=\theta_0(\delta)=\frac{1-\delta}{3}>0$,  $p_0=\frac{\rho}{2}$. For $ \hat\xi_{t}$ in \eqref{adasfcsa}, there is $\delta_0=\delta_0(\delta)>0$ such that for all $\rho\in(0,\rho_0)$ and $\theta\in(0,\theta_0)$, it holds 
\begin{align*}
	\PP \left({\sup_{0\leq t \leq T_\upsilon} \upsilon \left| \sum_{i=1}^{d} \int_{0}^{t}  g_{s-}^{i} d {\hat\xi}_{s}^{i} \right|\geq \upsilon^{\rho}}\right) \leq 2\exp\left(-\upsilon^{-p}\right),
\end{align*}
for all $p \in (0,p_0]$ and $0 < \upsilon \leq \upsilon_0$ with $\upsilon_0 = \upsilon(\rho)$, where ${\hat\xi}_{s}^{i}$ represents the i-th entry in ${\hat\xi}_{s}$.
\end{lemma}

\begin{lemma}[\citet{zhou2020generalizationdeep}]
\label{lemma4}
Under Assumptions \ref{assu:escape1}-\ref{assu:escape3} hold, assume $\delta \in (0,1), \rho_0 = \rho_0(\delta) = \frac{1-\delta}{16(1+c_1\kappa_1)} > 0, \theta_0=\theta_0(\delta)=\frac{1-\delta}{3}>0, p_0=\min(\frac{\hat\rho (1+c_1 \kappa_1)}{2}, p), \frac{1}{c_2} \ln\left(\frac{2\Delta}{\mu \upsilon^{\hat\rho}}\right)  \leq \upsilon^{-\theta_0}$ where $\kappa_1=\frac{c_2 l}{v_- |\tau_m-1|} $ and $c_2= \frac{2\mu\tau}{ \beta_1 v_+ +\mu\tau} \left(\beta_1 -\frac{\beta_2}{4} \right)$ in Adam-alike adaptive gradient algorithms. For all $\hat\rho\in(0,\rho_0)$, $p\in(0,p_0]$,   $0<\upsilon\leq \upsilon_0$ with $\upsilon_0 = \upsilon_0(\hat\rho)$, and $\theta_{0}= \hat\theta_{0}$, we have 
\begin{align}
	\sup_{\theta_{0}\in \boldsymbol{\Omega}} \PP\left({\sup_{0\leq t < \sigma_1} \norm{\theta_{t}- \hat\theta_{t}}_2 \geq 2\upsilon^{\hat\rho}}\right) \leq 2\exp(-\upsilon^{-\frac{p}{2}}), \label{small_dis_eq}
\end{align}
where the sequences $\theta_{t}$ and $\hat\theta_{t}$  are respectively produced by \eqref{sde_eq} and~\eqref{deterministicversion_adam_eq} in adaptive gradient method .
\end{lemma}
\subsubsection{Proof of Theorem~\ref{thm: escape_pf}}\label{sec: pf_for_escape}

\begin{proof}

The idea of this proof comes from \eqref{small_dis_eq} we showed in Lemma \ref{lemma4} where the sequence $\theta_t$ and $\hat\theta_t$ start from the same initialization. Based on Theorem \ref{thm: linearconvergenceadam}, we know that the sequence $\{\hat\theta_t\}$ from \eqref{deterministicversion_adam_eq} exponentially converges to the minimum $\theta^*$ of the local basin $\boldsymbol{\Omega}$. To escape the local basin $\boldsymbol{\Omega}$, we can either take small steps in the process $\zeta$ or large jumps $J_k$ in the process $\varepsilon$. However, \eqref{small_dis_eq} suggests that these small jumps might not be helpful for escaping the basin. And for big jumps, the escaping time $\Gamma$ of the sequence $\{\theta_t\}$ most likely occurs at the time $\sigma_1$ if the big jump $\upsilon J_1$ in the process $\varepsilon$ is large.
\newline The verification of our desired results can be divided into two separate parts, namely establishing upper bound and lower bound of $\mathbb{E}\left[{\exp{\left(-um(\Upsilon) \Theta(\upsilon^{-1}) \Gamma \right)}}\right]$ for any $u > -1$. Both of them can be established based on the following facts:
\begin{align}
    &\left\lvert \PP\left(R_\theta^{-1}\Sigma_\theta \upsilon J_k \not\in \boldsymbol{\Omega}^{\pm \upsilon^\gamma}, \norm{ \upsilon J_k }_2 \leq R\right) - \PP\left(R_{\theta^*}^{-1}\Sigma_{\theta^*} \upsilon J_k \not\in \boldsymbol{\Omega}^{\pm \upsilon^\gamma}, \norm{ \upsilon J_k }_2 \leq R\right) \right\rvert \leq \frac{\delta^\prime}{4} \cdot\frac{\Theta(\upsilon^{-1})}{\Theta(\upsilon^{-\delta})}, \nonumber  \\
    &\left\lvert \PP\left(R_\theta^{-1}\Sigma_\theta \upsilon J_k \not\in \boldsymbol{\Omega}, \norm{ \upsilon J_k }_2 \leq R\right) - \PP\left(R_{\theta^*}^{-1}\Sigma_{\theta^*} \upsilon J_k \not\in \boldsymbol{\Omega}, \norm{ \upsilon J_k}_2 \leq R\right) \right\rvert \leq \frac{\delta^\prime}{4}\cdot \frac{\Theta(\upsilon^{-1})}{\Theta(\upsilon^{-\delta})}, \nonumber \\
    &\PP\left(R_{\theta^*}^{-1}\Sigma_{\theta^*} \upsilon J_k \not\in \boldsymbol{\Omega}\right) - \PP\left(R_{\theta^*}^{-1}\Sigma_{\theta^*} \upsilon J_k \not\in \boldsymbol{\Omega}, \norm{\upsilon J_k}_2 \leq R\right) \leq \frac{\delta^\prime}{4} \cdot\frac{\Theta(\upsilon^{-1})}{\Theta(\upsilon^{-\delta})}. \label{three_eqs}
\end{align}
Specifically, for the upper bound of $\mathbb{E}\left[{\exp{\left(-um(\Upsilon) \Theta(\upsilon^{-1}) \Gamma \right)}}\right]$, we consider both the big jumps in the process $\varepsilon$ and small jumps in the process $\zeta$ which may escape the local minimum. Instead of estimating the escaping time $\Gamma$ from $\boldsymbol{\Omega}$, we first estimate the escaping time $\tilde \Xi$ from $\boldsymbol{\Omega}^{-\bar{\rho}}$. Here we define the inner part of $\boldsymbol{\Omega}$ as $\boldsymbol{\Omega}^{-\bar \rho} \coloneqq \{y \in \boldsymbol{\Omega} \, : \, \text{dis}(\partial \boldsymbol{\Omega} , y ) \geq \bar\rho \}$. Then by setting $\bar\rho \to 0$, we can use $\tilde \Xi$ for a decent estimation of $\Gamma$. We denote $\bar\rho = \upsilon^\gamma$ where $\gamma$ is a constant such that the results of Lemma \ref{lemma2}-\ref{lemma4} hold. So for the upper bound we mainly focus on $\tilde \Xi$ in the beginning and then transfer the results to $\Gamma$.
In the beginning, we can show that for any $u > -1$ it holds that,
\begin{equation*}
	\EE\left[{\exp{\left(-um(\Upsilon) \Theta(\upsilon^{-1}) \tilde \Xi \right)}}\right] \leq \sum_{k=1}^{+\infty} \EE \left[ e^{-um(\Upsilon) \Theta(\upsilon^{-1})  t_k} I\left\{\tilde \Xi=t_k\right\}  + Res_{k}\right],
\end{equation*}
	where 
\begin{align*}
	Res_{k}\leq 
	\begin{cases}
	\EE\left[e^{-um(\Upsilon) \Theta(\upsilon^{-1})t_k}I\left\{\tilde \Xi \in(t_{k-1}, t_k)\right\}\right],\quad &\text{if}\ u\in(-1, 0]\\
	\EE\left[e^{-um(\Upsilon) \Theta(\upsilon^{-1})t_{k-1}}I\left\{\tilde \Xi \in(t_{k-1}, t_k)\right\}\right],\quad &\text{if}\ u\in(0, +\infty).
	\end{cases}
\end{align*}
Then using the strong Markov property we can bound the first term $\EE \left[ e^{-um(\Upsilon) \Theta(\upsilon^{-1})  t_k} I\left\{\tilde \Xi=t_k\right\}\right]$ as
	\begin{align*}
    R_1 = \sum_{k=1}^{+\infty} \EE \left[e^{-um(\Upsilon) \Theta(\upsilon^{-1})t_k}I\left\{\Gamma=t_k\right\} \right]
	\leq & \frac{\alpha_\upsilon(1+\rho/3)}{1+u\alpha_\upsilon} \sum_{k=1}^{+\infty}\left(\frac{1-\alpha_\upsilon(1-\rho)}{1+u\alpha_\upsilon}\right)^{k-1} \\
	\leq & \frac{\alpha_\upsilon(1+\rho/3)}{1+u\alpha_\upsilon} \sum_{k=0}^{+\infty}\left(\frac{1-\alpha_\upsilon(1-\rho)}{1+u\alpha_\upsilon}\right)^{k-1} \\
	=& \frac{1+\rho/3}{1+u -\rho}.
	\end{align*}

On the other hand, for the lower bound of $\mathbb{E}\left[{\exp{\left(-um(\Upsilon) \Theta(\upsilon^{-1}) \Gamma \right)}}\right]$, we only consider the big jumps in the process $\varepsilon$ which could escape from the basin, and ignore the probability that the small jumps in the process $\zeta$ which may also lead to an escape from the local minimum $\theta^*$. Specifically, we can find a lower bound by discretization:
\begin{align*}
    \mathbb{E}\left[{\exp{\left(-um(\Upsilon) \Theta(\upsilon^{-1}) \Gamma \right)}}\right]
    \geq \sum_{k=1}^{+\infty} \mathbb{E}\left[{\exp{\left(-um(\Upsilon) \Theta(\upsilon^{-1}) t_k \right)}} I\{\Gamma = t_k\} \right].
\end{align*}
Then we can lower bound each term by three equations \eqref{three_eqs} we just listed here, which implies that for any $\theta_0 \in \boldsymbol{\Omega}^{-\upsilon^\gamma}$,
\begin{align*}
	\EE\left[e^{-um(\Upsilon)\Theta{\upsilon^{-1}}\Gamma} \right] \geq \frac{\alpha_\upsilon(1-\rho)}{1+u \alpha_\upsilon} \sum_{k=1}^{+\infty}\left(\frac{1- \alpha_\upsilon(1+\rho)}{1+u \alpha_\upsilon}\right)^{k-1} = \frac{1-\rho}{1+u +\rho},
\end{align*}
where $\rho \to 0$ as $\upsilon \to 0$. The proof is completed.
\end{proof}

\subsection{Proof of Proposition \ref{thm: generalize}}
\begin{proof}
Since we assumed the minimizer $\theta^* = {\bf 0}$ in the basin $\Omega$ which is usually small,we can employ second-order Taylor expansion to approximate $\Omega$ as a quadratic basin whose center is $\theta^*$. In other words, we can write
\begin{equation*}
    \Omega = \left\{y\in \RR^d ~\bigg |~ f(\theta^*) + \frac{1}{2} y^\top H(\theta^*) y \le h(\theta^*)\right\},
\end{equation*}
where $H(\theta^*)$ is the Hessian matrix at $\theta^*$ of function $f$ and $h(\theta^*)$ is the basin height. Then according to Definition~\ref{def: escaping}, we have
\begin{equation*}
    \Upsilon = \left\{y \in \mathbb{R}^d~\Big|~ y^{\top} \Sigma_{\theta^*} R_{\theta^*}^{-1} H(\theta^*) R_{\theta^*}^{-1} \Sigma_{\theta^*} y \geq h_f^*\right\}.
\end{equation*}
Here $R_{\theta^*} = \lim_{\theta_t \rightarrow \theta^*}  \text{diag}(\sqrt{v_t/(1-\beta_2^t)})$ is a matrix depending on the algorithm, $h_f^* = 2(h(\theta^*) - f(\theta^*))$ and $\Sigma_{\theta^*}$ is independent of the alogorithm, i.e. the same for Adam and AdaMomentum. Firstly, we will prove that $v_t^{(\textsc{AdaMomentum})} \geq v_t^{(\textsc{Adam})}$ when $t \to \infty$. To clarify the notation, we use $\theta_t, m_t, v_t, g_t$ to denote the symbols for Adam and $\tilde \theta_t, \tilde m_t,\tilde v_t,\tilde g_t$ for AdaMomentum, and $\zeta_t$ is the gradient noise. By using Lemma~\ref{thm: escape_time} and above results, we have $\theta_t \approx \tilde \theta_t \approx \theta^*$ before escaping when $t$ is large, and thus $v_t = \lim_{\theta_t \to \theta^*} [\nabla f(\theta_t) + \zeta_t]^2$ and $\tilde v_t = \lim_{\theta_t \to \theta^*} [\beta_1 \tilde m_{t-1} + (1-\beta_1)(\nabla f(\tilde\theta_t) +\zeta_t)]^2$. We will firstly show that $\EE(\tilde v_t) \geq \EE(v_t)$ when $t$ is large.
\begin{align*}
    \EE(v_t) &= \EE(\lim_{\theta_t \to \theta^*} [\nabla f(\theta_t) + \zeta_t]^2) \mathop{=}^{\textrm{(i)}} \lim_{\theta_t \to \theta^*}\EE([\nabla f(\theta_t) + \zeta_t]^2) \nonumber \\
    &= \lim_{\theta_t \to \theta^*} \left(\EE(\nabla f(\theta_t)^2) + \EE(2\nabla f(\theta_t)\zeta_t) + \EE(\zeta_t^2)\right) \\
    &\mathop{=}^{\textrm{(ii)}}  \EE(\lim_{\theta_t \to \theta^*} \nabla f(\theta_t)^2) + \lim_{\theta_t \to \theta^*} \EE( 2\nabla f(\theta_t)\zeta_t) + \lim_{\theta_t \to \theta^*} \EE(\zeta_t^2) \\
    &\mathop{=}^{\textrm{(iii)}} \lim_{\theta_t \to \theta^*} \EE(\zeta_t^2),
\end{align*}
where (i) and (ii) are due to the dominated convergence theorem (DCT) since we have that we know both $\norm{ \nabla f(\theta_t)}_2$ and $\norm{ \nabla f(\theta_t)+\zeta_t}_2 $ could be bounded by $H$ in Assumption \ref{asp: gradient-unbiased-noise-ind}. And (iii) is due to the fact that $\nabla f(\theta^*) = 0$ since function $f$ attains its minimum point at $\theta^*$, and $\zeta_t$ has zero mean, i.e. 
\begin{align*}
    \lim_{\theta_t \to \theta^*} \EE(\nabla f(\theta_t)\zeta_t) = 
    \lim_{\theta_t \to \theta^*} \EE(\nabla f(\theta_t))\EE(\zeta_t) = 0.
\end{align*}
And similarly we can prove that,
\begin{align*}
    \EE(\tilde v_t) &= \EE\left(\lim_{\theta_t \to \theta^*} [\beta_1 \tilde m_{t-1} + (1-\beta_1)(\nabla f(\tilde\theta_t) +\zeta_t)]^2\right) \\
    &= \lim_{\theta_t \to \theta^*} \left(\EE(\beta_1^2 \tilde m_{t-1}^2) + \EE((1-\beta_1)^2 (\nabla f(\tilde\theta_t) + \zeta_t)^2) + \EE(2\beta_1(1-\beta_1)\tilde m_{t-1} \nabla (f(\tilde\theta_t) +\zeta_t))\right) \\
    &\mathop{=}^{\textrm{(i)}} \beta_1^2 \, \lim_{\theta_t \to \theta^*} \EE(\tilde m_{t-1}^2) + (1-\beta_1)^2 \, \lim_{\theta_t \to \theta^*}\EE(\zeta_t^2),
\end{align*}
where we can get the equality (i) simply by the same argument with dominated convergence theorem we just used:
\begin{gather*}
    \lim_{\tilde \theta_t \to \theta^*} \EE(\nabla (f(\tilde\theta_t)^2) = \EE(\lim_{ \tilde \theta_t \to \theta^*} \nabla (f(\tilde\theta_t)^2) \mathop{=}^{\textrm{(i)}} 0, \\
    \lim_{\tilde \theta_t \to \theta^*} \EE(\nabla (f(\tilde\theta_t)  \zeta_t) = \EE(\lim_{\tilde \theta_t \to \theta^*} \nabla (f(\tilde\theta_t) \zeta_t) \mathop{=}^{\textrm{(ii)}} 0, \\
    \lim_{\tilde \theta_t \to \theta^*} \EE(\tilde m_{t-1} (\nabla f(\tilde\theta_t) + \zeta_t)) =
    \EE(\lim_{\tilde \theta_t \to \theta^*}\tilde m_{t-1} \nabla f(\tilde\theta_t)) + 
    \lim_{\tilde\theta_t \to \theta^*}\EE(\tilde m_{t-1}) \EE( \zeta_t) \mathop{=}^{\textrm{(iii)}} 0,
\end{gather*}
where we get the equality (i) and (ii) since the function $f(\tilde\theta_t)^2$ and $f(\tilde\theta_t) \zeta_t$ could be absolutely bounded by $H^2$. And the first term in equality (iii) is $0$ since we have $\norm{\tilde m_{t-1}}_2 \leq H$ by its definition and $\nabla f(\theta^*) = 0$, and the second term vanishes since the noise $\zeta_t$ has zero mean. Based on the Assumption \ref{asp:noise_decay}, we have 
\begin{align*}
    \EE(\tilde m_{t-1}^2) \geq \frac{2-\beta_1}{\beta_1} \EE(\zeta_t^2),
\end{align*}
which implies that $\EE(\tilde v_t) \geq \EE(v_t)$ when $t$ is large. It further indicates that $R_{\theta^*}^{(\textsc{AdaMomentum})} \ge R_{\theta^*}^{(\textsc{Adam})}$.
\newline We consider the volume of the complementary set 
\begin{equation*}
    \Upsilon^c = \left\{y \in \mathbb{R}^d~\Big|~ y^{\top} \Sigma_{\theta^*} R_{\theta^*}^{-1} H(\theta^*) R_{\theta^*}^{-1} \Sigma_{\theta^*} y < h_f^*\right\},
\end{equation*} 
which can be viewed as a $d$-dimensional ellipsoid. We can further decompose the symmetric matrix $M\coloneqq \Sigma_{\theta^*} R_{\theta^*}^{-1} H(\theta^*) R_{\theta^*}^{-1} \Sigma_{\theta^*}$ by SVD decomposition
\begin{equation*}
    M=U^\top A U,
\end{equation*}
where $U$ is an orthogonal matrix and $A$ is a diagonal matrix with nonnegative elements. Hence the transformation $y\rightarrow Uy$ is an orthogonal transformation which means the volume of $\Upsilon^c$ equals the volume of set
\begin{equation*}
    \left\{y' \in \mathbb{R}^d~\Big|~ y'^{\top} A y' < h_f^*\right\}.
\end{equation*} 
Considering the fact that the volume of a $d$-dimensional ellipsoid  centered at $\bf{0}$ $E_d(r) = \{(x_1, x_2, \cdots, x_n): \sum_{i=1}^d \frac{x_i^2}{R_i^2} \le 1\}$ is 
\begin{equation*}
    V(E_d(r) ) = \frac{\pi^{\frac{n}{2}}}{\Gamma(\frac{n}{2}+1)}\Pi_{i=1}^n R_i,
\end{equation*}
and the fact we just proved that $R_{\theta^*}^{(\textsc{AdaMomentum})} \geq R_{\theta^*}^{(\textsc{Adam})}$. Therefore we deduce the volume of  $\Upsilon^{(\textsc{AdaMomentum})}$ is smaller than that of $\Upsilon^{(\textsc{Adam})}$, which indicates that for Radon measure $m(\cdot)$ we have $m(\Upsilon^{(\textsc{AdaMomentum})}) \geq m(\Upsilon^{(\textsc{Adam})})$. Based on Lemma \ref{thm: escape_time}, we consequently have $\EE (\Gamma^{(\textsc{AdaMomentum})}) \ge \EE (\Gamma^{(\textsc{Adam})})$. 
\end{proof}

%% file: appendix_convex.tex
\section{Proofs in Section~\ref{sec: convergence}}\label{apd: proof-converge}
\subsection{Proof of the convergence results for the convex case}~\label{proof: convex}
\subsubsection{Proof of Theorem~\ref{thm: convex}}

\begin{proof}
  Firstly, according to the definition of AdaMomentum in Algorithm~\ref{alg: adamomentum}, by algebraic shrinking we have
    \begin{align*}
      \sum_{t=1}^T \frac{m_{t,i}^2}{\sqrt{tv_{t,i}}} &= \sum_{t=1}^{T-1} \frac{m_{t,i}^2}{\sqrt{tv_{t,i}}} + \frac{\left(\sum_{j=1}^T (1-\beta_{1,j})\Pi_{k=1}^{T-j} \beta_{1, T-k+1} g_{j,i}\right)^2}{\sqrt{T \left[\sum_{j=1}^T (1-\beta_2) \beta_2^{T-j} m_{j,i}^2 + \epsilon+ \sum_{j=1}^{T-1} \prod_{i=1}^j \beta_2^i \epsilon
      \right]}}\\
      &\leq  \sum_{t=1}^{T-1} \frac{m_{t,i}^2}{\sqrt{tv_{t,i}}} + \frac{\left(\sum_{j=1}^T (1-\beta_{1,j})\Pi_{k=1}^{T-j} \beta_{1, T-k+1} g_{j,i}\right)^2}{\sqrt{T \sum_{j=1}^T (1-\beta_2) \beta_2^{T-j} m_{j,i}^2}} \\
      & \le \sum_{t=1}^{T-1} \frac{m_{t,i}^2}{\sqrt{tv_{t,i}}} + \frac{(\sum_{j=1}^T \Pi_{k=1}^{T-j} \beta_{1, T-k+1})(\sum_{j=1}^T \Pi_{k=1}^{T-j} \beta_{1, T-k+1} g_{j,i}^2)}{\sqrt{T \sum_{j=1}^T (1-\beta_2) \beta_2^{T-j} m_{j,i}^2}} \\
      & \mathop{\le}^{\textrm{(i)}}  \sum_{t=1}^{T-1} \frac{m_{t,i}^2}{\sqrt{tv_{t,i}}} + \frac{(\sum_{j=1}^T \beta_1^{T-j})(\sum_{j=1}^T \beta_1^{T-j} g_{j,i}^2)}{\sqrt{T (1-\beta_2)\sum_{j=1}^T  \beta_2^{T-j} m_{j,i}^2}} \\
      & \le  \sum_{t=1}^{T-1} \frac{m_{t,i}^2}{\sqrt{tv_{t,i}}} + \frac{1}{1-\beta_1} \frac{\sum_{j=1}^T \beta_1^{T-j} g_{j,i}^2}{\sqrt{T (1-\beta_2)\sum_{j=1}^T  \beta_2^{T-j} m_{j,i}^2}} \\
      & = \sum_{t=1}^{T-1} \frac{m_{t,i}^2}{\sqrt{tv_{t,i}}} + \frac{1}{(1-\beta_1)\sqrt{T(1-\beta_2)}} \sum_{j=1}^T \frac{\beta_1^{T-j} g_{j,i}^2}{\sqrt{\sum_{j=1}^T \beta_2^{T-j} \left(\sum_{l=1}^j (1-\beta_{1,l}) \Pi_{k=1}^{j-l} \beta_{1, j-k+1} g_{l,i}\right)^2}} \\
      & \le \sum_{t=1}^{T-1} \frac{m_{t,i}^2}{\sqrt{tv_{t,i}}} + \frac{1}{(1-\beta_1)\sqrt{T(1-\beta_2)}} \sum_{j=1}^T \frac{\beta_1^{T-j} g_{j,i}^2}{\sqrt{\sum_{j=1}^T \beta_2^{T-j} \left( (1-\beta_{1,j})  g_{j,l}\right)^2}} \\
      & \le \sum_{t=1}^{T-1} \frac{m_{t,i}^2}{\sqrt{tv_{t,i}}} + \frac{1}{(1-\beta_1)\sqrt{T(1-\beta_2)}} \sum_{j=1}^T \frac{\beta_1^{T-j} g_{j,i}^2}{\sqrt{\beta_2^{T-j} (1-\beta_{1,j})^2 g_{j,i}^2}} \\
      & \mathop{\leq}^{\textrm{(ii)}} \sum_{t=1}^{T-1} \frac{m_{t,i}^2}{\sqrt{tv_{t,i}}} + \frac{1}{(1-\beta_1)^2\sqrt{T(1-\beta_2)}} \sum_{j=1}^T \gamma^{T-j} g_{j,i},
    \end{align*}
    where (i) arises from $\beta_{1,t} \le \beta_1$, and (ii) comes from the definition that $\gamma = \frac{\beta_1}{\sqrt{\beta_2}}$. Then by induction, we have
    \begin{align*}
      \sum_{t=1}^T \frac{m_{t,i}^2}{\sqrt{tv_{t,i}}} &\le \sum_{t=1}^T  \frac{1}{(1-\beta_1)^2 \sqrt{t(1-\beta_2)}} \sum_{j=1}^t \gamma^{t-j} g_{j,i} \\
      & \le \frac{1}{(1-\beta_1)^2 \sqrt{1-\beta_2}} \sum_{t=1}^T \frac{1}{\sqrt{t}} \sum_{j=1}^t \gamma^{t-j} g_{j,i} \\
      & \mathop{\le}^{(i)} \frac{1}{(1-\beta_1)^2 \sqrt{1-\beta_2}} \sum_{t=1}^T g_{t,i} \sum_{j=t}^T \frac{\gamma^{j-t}}{\sqrt{j}} \\
      & \le \frac{1}{(1-\beta_1)^2 \sqrt{1-\beta_2}} \sum_{t=1}^T g_{t,i} \cdot\frac{1}{(1-\gamma)\sqrt{t}} \\
      & \le \frac{1}{(1-\beta_1)^2 (1-\gamma)\sqrt{1-\beta_2}} \sum_{t=1}^T \frac{g_{t,i}}{\sqrt{t}} \\
      & \mathop{\le}^{\textrm{(ii)}} \frac{1}{(1-\beta_1)^2 (1-\gamma)\sqrt{1-\beta_2}} \norm{g_{1:T, i}}_2 \sqrt{\sum_{t=1}^T \frac{1}{t}} \\
      & \mathop{\le}^{\textrm{(iii)}} \frac{\sqrt{1+\log T}}{(1-\beta_1)^2 (1-\gamma)\sqrt{1-\beta_2}} \norm{g_{1:T, i}}_2,
    \end{align*}
    where (i) exchangings the indices of summing, (ii) employs Cauchy-Schwarz Inequality and (iii) comes from the following bound on harmonic sum: 
    \begin{equation*}
      \sum_{t=1}^T \frac{1}{t} \le 1+ \log T.
    \end{equation*}
    Due to convexity of $f_t$, we get 
    \begin{align} \label{eq: convex}
      f_t(\theta_t) - f_t(\theta^*) &\le g_t^\top (\theta_t - \theta^*) \nonumber \\
      & = \sum_{i=1}^d g_{t,i} (\theta_{t,i} - \theta_{,i}^*).
    \end{align}
    According to the updating rule, we have
    \begin{align}\label{eq: update-rule}
      \theta_{t+1} &= \theta_t - \alpha_t \frac{m_t}{\sqrt{v_t}} \nonumber\\
      & = \theta_t - \alpha_t \left (\frac{\beta_{1,t}}{\sqrt{v_{t}}} m_{t-1} + \frac{1-\beta_{1,t}}{\sqrt{v_{t}}}g_{t}\right).
    \end{align}
    Substracting $\theta^*$, squaring both sides and considering only the $i$-th element in vectors, we obtain 
    \begin{equation*}
      (\theta_{t+1,i} - \theta_{,i}^*)^2 = (\theta_{t,i} - \theta_{,i}^*)^2 - 2\alpha_t \left (\frac{\beta_{1,t}}{\sqrt{v_{t,i}}} m_{t-1,i} + \frac{1-\beta_{1,t}}{\sqrt{v_{t,i}}}g_{t,i}\right )(\theta_{t,i} - \theta_{,i}^*) + \alpha_t^2 \left (\frac{m_{t,i}}{\sqrt{v_{t,i}}}\right )^2.
    \end{equation*}
    By rearranging the terms, we have
    \begin{equation*}
      2\alpha_t \frac{1-\beta_{1,t}}{\sqrt{v_{t,i}}} g_{t,i} (\theta_{t,i} - \theta_{,i}^*) = (\theta_{t,i} - \theta_{,i}^*)^2 - (\theta_{t+1,i} - \theta_{,i}^*)^2 - 2\alpha_t \cdot \frac{\beta_{1,t}}{\sqrt{v_{t,i}}}\cdot m_{t-1,i} (\theta_{t,i} - \theta_{,i}^*) + \alpha_t^2 \left (\frac{m_{t,i}}{\sqrt{v_{t,i}}}\right )^2.
    \end{equation*}
    Further we have
    \begin{align}
      g_{t,i} (\theta_{t,i} - \theta_{,i}^*) =& \frac{\sqrt{v_{t,i}}}{2 \alpha_t (1-\beta_{1,t})} [(\theta_{t,i} - \theta_{,i}^*)^2 - (\theta_{t+1, i} - \theta_{,i}^*)^2] +  \frac{\alpha_t \sqrt{v_{t,i}}}{2(1-\beta_{1,t})} \left (\frac{m_{t,i}}{\sqrt{v_{t,i}}}\right )^2  \nonumber\\
      &+ \frac{\beta_{1,t}}{1-\beta_{1,t}}(\theta_{,i}^* - \theta_{t,i})m_{t-1,i} \nonumber\\
      = & \frac{\sqrt{v_{t,i}}}{2 \alpha_t (1-\beta_{1,t})} [(\theta_{t,i} - \theta_{,i}^*)^2 - (\theta_{t+1, i} - \theta_{,i}^*)^2] +  \frac{\alpha_t \sqrt{v_{t,i}}}{2(1-\beta_{1,t})} \left (\frac{m_{t,i}}{\sqrt{v_{t,i}}}\right )^2 \nonumber\\
      & +\frac{\beta_{1,t}}{1-\beta_{1,t}} \cdot\frac{v_{t,i}^{\frac{1}{4}}}{\sqrt{\alpha_{t}}}\cdot (\theta_{,i}^* - \theta_{t,i})\cdot \sqrt{\alpha_{t}}\cdot \frac{m_{t-1,i}}{v_{t,i}^{\frac{1}{4}}} \nonumber\\
      \le & \frac{\sqrt{v_{t,i}}}{2\alpha_t (1-\beta_1)} [(\theta_{t,i} - \theta_{,i}^*)^2 - (\theta_{t+1, i} - \theta_{,i}^*)^2] + \frac{\alpha}{2(1-\beta_1)}\cdot\frac{m_{t,i}^2}{\sqrt{t v_{t,i}}} \label{eq: before-sum-last-but-one}\\
      & + \frac{\beta_{1,t}}{2\alpha_{t} (1-\beta_{1,t})} (\theta_{,i}^* - \theta_{t,i})^2 \sqrt{v_{t, i}} + \frac{\beta_1 \alpha}{2(1-\beta_1)} \cdot\frac{m_{t-1,i}^2}{\sqrt{t v_{t,i}}} \label{eq: before-sum},
    \end{align}
    where \eqref{eq: before-sum} bounds the last term of \eqref{eq: before-sum-last-but-one} by Cauchy-Schwarz Inequality and plugs in the value of $\alpha_t$.
    Plugging \eqref{eq: before-sum} into \eqref{eq: update-rule} and summing from $t=1$ to $T$, we obtain
    \begin{align}
      R(T) = & \sum_{t=1}^T \sum_{i=1}^d  g_{t,i} (\theta_{t,i} - \theta_{,i}^*) \nonumber\\
        \le & \sum_{t=1}^T \sum_{i=1}^d \frac{\sqrt{v_{t,i}}}{2\alpha_t (1-\beta_1)} [(\theta_{t,i} - \theta_{,i}^*)^2 - (\theta_{t+1, i} - \theta_{,i}^*)^2] + \sum_{t=1}^T \sum_{i=1}^d \frac{\alpha}{2(1-\beta_1)}\cdot\frac{m_{t,i}^2}{\sqrt{t v_{t,i}}}  \nonumber\\
        & + \sum_{t=1}^T \sum_{i=1}^d \frac{\beta_{1,t}}{2\alpha_t (1-\beta_{1,t})} (\theta_{,i}^* - \theta_{t,i})^2 \sqrt{v_{t, i}} + \sum_{t=1}^T \sum_{i=1}^d \frac{\beta_1 \alpha}{2(1-\beta_1)} \cdot\frac{m_{t-1,i}^2}{\sqrt{t v_{t,i}}} \nonumber\\
        \le & \sum_{i=1}^d \frac{\sqrt{v_{1,i}}}{2\alpha_1 (1-\beta_1)} (\theta_{1,i} - \theta_{,i}^*)^2 + \frac{1}{2(1-\beta_1)}\sum_{t=2}^T \sum_{i=1}^d (\theta_{t,i} - \theta_{,i}^*)^2 \left(\frac{\sqrt{v_{t,i}}}{\alpha_t} - \frac{\sqrt{v_{t-1,i}}}{\alpha_{t-1}}\right) \label{eq: R-bound-before-D-last-but-one} \\
        & + \sum_{t=1}^T \sum_{i=1}^d \frac{\beta_{1,t}}{2\alpha_t(1-\beta_1)} (\theta_{,i}^* - \theta_{t,i})^2 \sqrt{v_{t,i}} + \sum_{t=1}^T \sum_{i=1}^d \frac{\alpha}{1-\beta_1}\cdot\frac{m_{t,i}^2}{\sqrt{t v_{t,i}}} \label{eq: R-bound-before-D},
    \end{align}
    where \eqref{eq: R-bound-before-D} rearranges the first term of \eqref{eq: R-bound-before-D-last-but-one}.
    Finally utilizing the assumptions in Theorem~\ref{thm: convex}, we get
    \begin{align}\label{eq: thm-1}
      R(T) \le& \sum_{i=1}^d \frac{\sqrt{v_{1,i}}}{2\alpha_1 (1-\beta_1)} D_{\infty}^2 +  \frac{1}{2(1-\beta_1)}\sum_{t=2}^T \sum_{i=1}^d D_{\infty}^2 \left(\frac{\sqrt{v_{t,i}}}{\alpha_t} - \frac{\sqrt{v_{t-1,i}}}{\alpha_{t-1}}\right) \nonumber\\
      & + \frac{D_{\infty}^2}{2(1-\beta_1)} \sum_{t=1}^T \sum_{i=1}^d \frac{\beta_{1,t} v_{t,i}^{\frac{1}{2}}}{\alpha_t} + \sum_{i=1}^d \frac{\alpha \sqrt{1+\log T}}{(1-\beta_1)^3 (1-\gamma)\sqrt{1-\beta_2}} \norm{g_{1:T, i}}_2 \nonumber\\
      = & \sum_{i=1}^d \frac{\sqrt{v_{T,i}}}{2\alpha_T (1-\beta_1)} D_{\infty}^2 + \frac{D_{\infty}^2}{2(1-\beta_1)} \sum_{t=1}^T \sum_{i=1}^d \frac{\beta_{1,t} v_{t,i}^{\frac{1}{2}}}{\alpha_t} \nonumber\\
      &+ \sum_{i=1}^d \frac{\alpha \sqrt{1+\log T}}{(1-\beta_1)^3 (1-\gamma)\sqrt{1-\beta_2}} \norm{g_{1:T, i}}_2,
    \end{align}
    which is our desired result.
\end{proof}

\subsubsection{Proof of Corollary~\ref{cor: convex}}
\begin{proof}
  Plugging $\alpha_t = \frac{\alpha}{\sqrt{t}}$ and $\beta_{1,t} = \beta_1 \lambda^t$ into \eqref{eq: thm-1}, we get
  \begin{align}\label{eq: RT-bound}
    R(T) \le &  \frac{D_{\infty}^2\sqrt{T}}{2\alpha (1-\beta_1)} \sum_{i=1}^d\sqrt{v_{T,i}}  + \frac{D_{\infty}^2}{2\alpha(1-\beta_1)} \sum_{t=1}^T \sum_{i=1}^d \beta_{1}\lambda^t \sqrt{tv_{t,i}} \nonumber\\
    &+ \sum_{i=1}^d \frac{\alpha \sqrt{1+\log T}}{(1-\beta_1)^3 (1-\gamma)\sqrt{1-\beta_2}} \norm{g_{1:T, i}}_2. 
  \end{align}
  Next, we employ Mathematical Induction to prove that $v_t,i\le G_{\infty}$ for any $0 \le t \le T, 1 \le i \le d$. $\forall i$, we have $m_{0,i}^2 = 0 \le G_{\infty}^2$. Suppose $m_{t-1,i} \le G_{\infty}$, we have
  \begin{align*}
    m_{t,i}^2 &= \left(\beta_{1,t} m_{t-1,i} + (1-\beta_{1,t}) g_{t,i}\right)^2 \\
    & \mathop{\le}^{\textrm{(i)}} \beta_{1,t} m_{t-1,i}^2 + (1-\beta_{1,t}) g_{t,i}^2 \\
    & \le \beta_{1,t} G_{\infty}^2 + (1-\beta_{1,t}) G_{\infty}^2 = G_{\infty}^2,
  \end{align*}
  where (i) comes from the convexity of function $f=x^2$. Hence by induction, we have $m_{t,i}^2 \le G_{\infty}^2$ for all $0 \le t \le T$. Furthermore, $\forall i,$ we have $v_{0,i} = 0 \le G_{\infty}^2$. Suppose $v_{t-1,i} \le G_{\infty}^2 + (1-\beta_2^{t-1})\epsilon/(1-\beta_2)$, we have 
  \begin{align*}
    v_{t,i} &= \beta_2 v_{t-1,i} + (1-\beta_2) m_{t,i}^2 + \epsilon \\
    & \le \beta_2 G_{\infty}^2 + (1-\beta_2) G_{\infty}^2 + \left(\frac{\beta_2-\beta_2^t}{1-\beta_2}+1\right)\epsilon = G_{\infty}^2+ \frac{1-\beta_2^t}{1-\beta_2} \epsilon.
  \end{align*} 
  Therefore, by induction, we have $v_{t,i} \le G_{\infty}^2 + (1-\beta_2^{t})\epsilon/(1-\beta_2) \leq G_{\infty}^2 +  \epsilon/(1-\beta_2), \forall i, t$. Combining this with the fact that $\sum_{i=1}^d \norm{g_{1:T,i}}_2 \le dG_{\infty}\sqrt{T}$ and \eqref{eq: RT-bound}, we obtain
  \begin{equation}\label{eq: R(T)-bound-lastbutone}
    R(T) \le \frac{d(G_{\infty}+\sqrt{\frac{\epsilon}{1-\beta_2}})D_{\infty}^2\sqrt{T}}{2\alpha (1-\beta_1)} + \frac{d(G_{\infty}+\sqrt{\frac{\epsilon}{1-\beta_2}})D_{\infty}^2 \beta_{1}}{2\alpha(1-\beta_1)} \sum_{t=1}^T \lambda^t \sqrt{t} + \frac{dG_{\infty}\alpha \sqrt{1+\log T}}{(1-\beta_1)^3 (1-\gamma) \sqrt{(1-\beta_2)T}}.
  \end{equation}
  For $\sum_{t=1}^T \lambda^t \sqrt{t}$, we apply arithmetic geometric series upper bound:
  \begin{equation}\label{eq: geometric-bound}
    \sum_{t=1}^T \lambda^t \sqrt{t} \le \sum_{t=1}^T t\lambda^t \le \frac{1}{(1-\lambda)^2}.
  \end{equation}
  Plugging \eqref{eq: geometric-bound} into \eqref{eq: R(T)-bound-lastbutone} and dividing both sides by $T$, we obtain
  \begin{equation*}
    \frac{R(T)}{T} \le \frac{d(G_{\infty}+\sqrt{\frac{\epsilon}{1-\beta_2}})\alpha \sqrt{1+\log T}}{(1-\beta_1)^3 (1-\gamma) \sqrt{(1-\beta_2)T}}+\frac{d D_{\infty}^2 (G_{\infty}+\sqrt{\frac{\epsilon}{1-\beta_2}})}{2\alpha (1-\beta_1)\sqrt{T}} + \frac{d D_{\infty}^2 G_{\infty}\beta_1}{2\alpha (1-\beta_1)(1-\lambda)^2 T},
  \end{equation*}
  which concludes the proof.
\end{proof}

%% file: appendix_nonconvex.tex
\subsection{Proof of the convergence results for the non-convex case}~\label{proof: nonconvex}
\subsubsection{Useful Lemma}
\begin{lemma}\label{lem: tool_nonconvex}(\citet{wang2017accelerating,guo2021stochastic}) Consider a moving average sequence $m_{t+1} = \beta_{1,t} m_t + (1-\beta_t) g_{t+1}$ for tracking $\nabla f(\theta_t)$, where $\mathbb{E}(g_{t+1}) = \nabla f(\theta_t)$ and $f$ is an L-Lipschits continuous mapping. Then we have
$$ \mathbb{E}_t(\norm{m_{t+1} - \nabla f(\theta_t)}_2^2) \leq \beta_{1,t} \norm{m_{t} - \nabla f(\theta_{t-1})}_2^2 + 2(1-\beta_{1,t})^2 \mathbb{E}_t(\norm{g_{t+1}-\nabla f(\theta_t)}_2^2) + \frac{L^2}{1-\beta_{1,t}} \norm{\theta_t - \theta_{t-1}}_2^2.
$$
\end{lemma} 
Based on the above Lemma \ref{lem: tool_nonconvex}, we could derive the following convergence result in Theorem \ref{thm: nonconvex}.
\subsubsection{Proof of Theorem \ref{thm: nonconvex}}

We denote $\Delta_t = \norm{m_{t+1} - \nabla f(\theta_2)}_2^2$, and by applying Lemma \ref{lem: tool_nonconvex} we can get:
\begin{align}
    \mathbb{E}_t(\Delta_{t+1}) \leq \beta_{1,t+1} \Delta_t + 2(1-\beta_{1,t+1})^2 \mathbb{E}_t(\norm{g_{t+2} - \nabla f (\theta_{t+1})}_2^2) + \frac{L^2}{1-\beta_{1,t+1}} \norm{\theta_{t+1} - \theta_t}_2^2. \label{eq:above}
\end{align}
Based on some simple calculation, we can verify that $\sum_{i=0}^{t-1} \beta^i \epsilon$ elementwisely, which implies that $1/\sqrt{v_t} \leq b_{u,t}$ holds for all $t \in [T]$. On the other hand, since we have $m_{t+1} = \beta_{1,t} m_{t-1} + (1-\beta_{1,t}) g_t$ with the condition $\norm{g_t}_\infty \leq G$ for all $t \in [T]$. Therefore, we can deduce that
$$\norm{m_t}_\infty \leq \beta_{1,t} \norm{m_{t-1}}_\infty + (1-\beta_{1,t}) G \leq \beta \norm{m_{t-1}}_\infty + (1-\beta) G, \quad \; m_0 = 0,$$
which implies that $\norm{m_t}_\infty \leq G(1-\beta^t)$ after some simple calculation, and hence we have $m_t^2 \leq G^2(1-\beta^T)^2$ elementwise. Next, since we have $v_{t+1} = \beta_{2} v_{t-1} + (1-\beta_{2}) m_t^2 + \epsilon$, we can similarly get
$$\norm{v_t}_\infty \leq \beta_{2} \norm{v_{t-1}}_\infty + (1-\beta_{2})\left(G^2(1-\beta_{1,1}^T) + \frac{\epsilon}{1-\beta_{2}}\right), \quad \; v_0 = 0,$$
which implies that $\norm{v_t}_\infty \leq \left(G^2(1-\beta^T) + \frac{\epsilon}{1-\beta_{2}}\right)(1-\beta_2^t)$ and hence $1/\sqrt{v_t} \geq b_{l,t}$. After some simple simplification of Eqn. \eqref{eq:above}, we have
\begin{align}
\mathbb{E}_t(\sum_{t=0}^T (1-\beta_{1,t+1}) \Delta_t) 
&\leq \mathbb{E}\left[ \sum_{t=0}^T (\Delta_t - \Delta_{t-1}) + \sum_{t=0}^T 2 \sigma^2 (1-\beta_{1,t+1})^2 + \sum_{t=0}^T \frac{L^2}{1-\beta_{1,t+1}} \norm{\theta_{t+1} - \theta_t}_2^2
\right] \nonumber\\
&\mathop{=}^{\textrm{(i)}} \mathbb{E}\left[ \sum_{t=0}^T (\Delta_t - \Delta_{t-1}) + \sum_{t=0}^T 2 \sigma^2 (1-\beta_{1,t+1})^2 + \sum_{t=0}^T \frac{L^2\alpha_t^2 b_{u,t+1}^2}{1-\beta_{1,t+1}} \norm{m_{t+1}}_2^2
\right], \label{eq:noncov}
\end{align}
where (i) comes from the Lipschitz property of $\nabla f$. On the other hand, since $f$ has Lipschitz gradient, we have:
\begin{align}
    f(\theta_{t+1}) &\leq f(\theta_t) + \nabla f(\theta_t)^\top (\theta_{t+1} - \theta_t) + \frac{L}{2} \norm{\theta_{t+1} - \theta_t}_2^2 \nonumber \\
    &= f(\theta_t) - \nabla f(\theta_t)^\top (\frac{\alpha_t}{\sqrt{v_t}} m_{t+1}) + \frac{L}{2} \norm{\frac{\alpha_t}{\sqrt{v_t}} m_{t+1}}_2^2 \nonumber \\
    &= f(\theta_t) + \frac{\alpha_t}{2\sqrt{v_t}} \norm{\nabla f(\theta_t) - m_{t+1}}_2^2 + \frac{L}{2} \norm{\frac{\alpha_t}{\sqrt{v_t}} m_{t+1}}_2^2 - \frac{\alpha_t}{2\sqrt{v_t}} \norm{\nabla f(\theta_t) }_2^2 - \frac{\alpha_t}{2\sqrt{v_t}} \norm{m_{t+1}}_2^2 \nonumber \\
    &\leq f(\theta_t) + \frac{\alpha_t b_{u,t}}{2} \Delta_t + \frac{L\alpha_t^2b_{u,t}^2 - \alpha_t b_{l,t}}{2} \norm{m_{t+1}}_2^2 - \frac{\alpha_t b_{l,t}}{2} \norm{\nabla f(\theta_t) }_2^2. \label{eq:noncov_lemma}
\end{align}
Since we know that $T_0 \lesssim \frac{1}{\alpha_T}$, then we know the overall loss of the first $T_0$ terms would be $\mathbb{E}(\sum_{t=1}^{T_0} \norm{\nabla f(\theta)}_2^2) \lesssim 1/\alpha_T$, and hence 
\begin{align}
\mathbb{E}\left(\frac{1}{T+1}\sum_{t=1}^{T_0} \norm{\nabla f(\theta)}_2^2\right) \lesssim \frac{1}{\alpha_T(T+1)}. \label{eq:noncov_first}
\end{align}
For the other case when $t > T_0$, without loss of generality we can assume that $T_0=0$ for the above argument. We denote $A = \sqrt{\frac{b_{l,T}}{2L^2b_{u,1}^3}}$ and $\theta^* =\arg\min_\theta f(\theta)$. From Eqn. \eqref{eq:noncov_lemma}, we have
\begin{align*}
    \mathbb{E}\left(\sum_{t=0}^T \frac{\alpha_t b_{l,t}}{2} \norm{\nabla f(\theta_t)}_2^2 \right) &\leq \mathbb{E}\left[\sum_{t=0}^T (f(\theta_t)-f(\theta_{t+1}))+ \sum_{t=0}^T \frac{\alpha_t b_{u,t}}{2} \Delta_t +\sum_{t=0}^T \frac{(L \alpha_t^2 b_{u,t}^2 - \alpha b_{l,t})}{2} \norm{m_{t+1}}_2^2\right] \\ 
    &\leq f(\theta_0) - f(\theta^*) + \mathbb{E}\left(\sum_{t=0}^T \frac{(L \alpha_t^2 b_{u,t}^2 - \alpha b_{l,t})}{2} \norm{m_{t+1}}_2^2\right) + \mathbb{E}\left(\sum_{t=0}^T \frac{\alpha_t b_{u,t}}{2(1-\beta_{1,t+1})} (1-\beta_{1,t+1}) \Delta_t\right) \\
    &\mathop{\leq}^{\textrm{(i)}} f(\theta_0) - f(\theta^*) + \mathbb{E}\left(\sum_{t=0}^T \frac{(L \alpha_t^2 b_{u,t}^2 - \alpha b_{l,t})}{2} \norm{m_{t+1}}_2^2\right) + \frac{A b_{u,1}}{2}\mathbb{E}\left(\sum_{t=0}^T  (1-\beta_{1,t+1}) \Delta_t\right) \\
    &\mathop{\leq}^{\textrm{(ii)}} f(\theta_0) - f(\theta^*) + \mathbb{E}\left(\sum_{t=0}^T \frac{(L \alpha_t^2 b_{u,t}^2 - \alpha b_{l,t})}{2} \norm{m_{t+1}}_2^2\right)  \\
    & \quad \; +\frac{A b_{u,1}}{2}\mathbb{E}\left[
    \Delta_0 + \sum_{t=0}^T 2(1-\beta_{1,t+1})^2 \sigma^2+ \sum_{t=0}^T \frac{L^2\alpha_t^2b_{u,t+1}^2}{1-\beta_{1,t+1}} \norm{m_{t+1}}_2^2\right] \\
    &\mathop{\leq}^{\textrm{(iii)}}  f(\theta_0) - f(\theta^*)+ \frac{Ab_{u,1}}{2}(\sigma^2+\norm{\nabla f(\theta_0)}_2^2) + Ab_{u,1} \sigma^2 \sum_{t=0}^T (1-\beta_{1,t+1})^2 \\
    & \quad \; + \mathbb{E}\left[\sum_{t=0}^T \left(\frac{AL^2b_{u,1} \alpha_t^2 b_{u,t+1}^2}{2(1-\beta_{1,t+1})} + \frac{L\alpha_t^2b_{u,t}^2 - \alpha_t b_{l,t}}{2} \right) \norm{m_{t+1}}_2^2 \right] \\
    &\mathop{\leq}^{\textrm{(iv)}}  f(\theta_0) - f(\theta^*)+ \frac{Ab_{u,1}}{2}(\sigma^2+\norm{\nabla f(\theta_0)}_2^2) + Ab_{u,1} \sigma^2 \sum_{t=0}^T (1-\beta_{1,t+1})^2,
\end{align*}
where \textrm{(i)} comes from the fact that $\alpha_t \leq (1-\beta_{1,t+1}) A$ based on the conditions in Theorem \ref{thm: nonconvex}; \textrm{(ii)} could be obtained after we apply Eqn. \eqref{eq:noncov} to the summation; \textrm{(iii)} is due to the fact that 
\begin{align*}
\mathbb{E}(\Delta_0) &= \mathbb{E}(\norm{(1-\beta_{1,1})(g_1 - \nabla f(\theta_0)) - \beta_{1,1}\nabla f(\theta_0) )}_2^2) \\
&=(1-\beta_{1,1})^2 \mathbb{E}(\norm{g_1 - \nabla f(\theta_0)}_2^2) + \beta_{1,1}^2 \mathbb{E}{\norm{\nabla f(\theta_0)}_2^2} \leq \sigma^2 +\norm{\nabla f(\theta_0)}_2^2. 
\end{align*}
And we can deduce \textrm{(iv)} by using the assumptions in Theorem \ref{thm: nonconvex}
\begin{gather*}
    \frac{AL^2b_{u,1} \alpha_t^2 b_{u,t+1}^2}{2(1-\beta_{1,t+1})} \leq \frac{A^2L^2b_{u,1}  b_{u,t+1}^2}{2} \leq \frac{b_{l,t}}{4}, \quad \frac{L \alpha_t b_{u,t}^2}{2} \leq \frac{b_{l,t}}{4}.
\end{gather*}
Therefore, we have
\begin{align*}
     \mathbb{E}\left(\sum_{t=0}^T \frac{\alpha_T b_{l,T}}{2} \norm{\nabla f(\theta_t)}_2^2 \right) &\leq \mathbb{E}\left(\sum_{t=0}^T \frac{\alpha_t b_{l,t}}{2} \norm{\nabla f(\theta_t)}_2^2 \right) \\
     &\leq f(\theta_0) - f(\theta^*)+ \frac{Ab_{u,1}}{2}(\sigma^2+\norm{\nabla f(\theta_0)}_2^2) + Ab_{u,1} \sigma^2 \sum_{t=0}^T (1-\beta_{1,t+1})^2.
     \end{align*}
As a consequence, we can deduce that
\begin{align}
\mathbb{E}\left(\sum_{t=0}^T \frac{1}{T+1} \norm{\nabla f(\theta_t)}_2^2 \right) &\leq \dfrac{\frac{2(f(\theta_0)-f(\theta^*))}{b_{l,T}} + \sqrt{\frac{1}{2L^2b_{u,1} b_{l,T}}}(\sigma^2 + \norm{\nabla f(\theta_0)}_2^2)}{\alpha_T (T+1)} + \sqrt{\frac{2}{Lb_{u,1} b_{l,T}}} \frac{\sigma^2 \sum_{t=0}^T (1-\beta_{1,t+1})^2}{\alpha_T (T+1)} \label{eq:noncov_res}\\
& \coloneqq \frac{1}{\alpha_T(T+1)} (Q_1 + Q_2 \eta(T)),  \nonumber
\end{align} 
where
$$Q_1 = \frac{2(f(\theta_0)-f(\theta^*))}{b_{l,T}} + \sqrt{\frac{1}{2L^2b_{u,1} b_{l,T}}}(\sigma^2 + \norm{\nabla f(\theta_0)}_2^2), \quad Q_2 = \sqrt{\frac{2}{Lb_{u,1} b_{l,T}}}\sigma^2.$$
\hfill  \qedsymbol
\subsubsection{Proof of Corollary \ref{cor: nonconvex}}
Without loss of generality we choose $1-\beta_{1,t} = \beta / \sqrt{t} \text{ and } \alpha_t = \alpha/ \sqrt{t}, \forall t \in [T]$ for some constants $\alpha,\beta$ with all conditions in Theorem \ref{thm: nonconvex} hold, we have
\begin{align*}
    T\alpha_T = \alpha \sqrt{T}, \quad \eta(T) = \sum_{t=1}^T (1-\beta_{1,t})^2 = \beta^2 \sum_{t=1}^T \frac{1}{t} \leq \beta^2(1+\log(T)).
\end{align*}
After combining this with~\eqref{eq:noncov_res} and making some rearrangement, we have:
\begin{align}
\mathbb{E}\left(\sum_{t=0}^T \frac{1}{T+1} \norm{\nabla f(\theta_t)}_2^2 \right) &\leq \dfrac{\frac{2(f(\theta_0)-f(\theta^*))}{b_{l,T}} + \sqrt{\frac{1}{2L^2b_{u,1} b_{l,T}}}(\sigma^2 + \norm{\nabla f(\theta_0)}_2^2)}{\alpha \sqrt{T}} + \sqrt{\frac{2}{L b_{u,1} b_{l,T}}} \frac{\sigma^2 \beta^2(1+\log(T))}{\alpha \sqrt{T}} \nonumber\\
& \coloneqq \frac{1}{\alpha_T(T+1)} (Q_1^* + Q_2^* \eta(T)),  \nonumber
\end{align} 
where
$$Q_1^* = \frac{2(f(\theta_0)-f(\theta^*))}{b_{l,T} \alpha} + \sqrt{\frac{1}{2L^2 b_{u,1} b_{l,T} }}\frac{(\sigma^2 + \norm{\nabla f(\theta_0)}_2^2)}{\alpha} + \sqrt{\frac{2}{Lb_{u,1} b_{l,T}}}\frac{\sigma^2 \beta^2}{\alpha}, \quad Q_2^* = \sqrt{\frac{2}{Lb_{u,1} b_{l,T}}}\frac{\sigma^2 \beta^2}{\alpha}.$$
\hfill  \qedsymbol

%% file: appendix_exp.tex
\section{Additional Experimental Details}\label{apd: details_exp}

\subsection{Hyperparameter tuning rule} For hyperparameter tuning, we perform extensive and careful grid search to choose the best hyperparameters for all the baseline algorithms. 

\paragraph{CNN for Image Classification} For SGDM, we set the momentum as $0.9$ which is the default choice~\cite{he2016deep,huang2017densely} and search the learning rate between $0.1$ and $10^{-5}$ in the log-grid. For all the adaptive gradient methods, we fix $\beta_1=0.9$ and $\beta_2=0.999$ and search the learning rate between $0.1$ and $1e^{-5}$ in the log-grid, $\epsilon$ between $1e^{-5}$ and $1e^{-16}$ in the log-grid. For all optimizers we grid search weight decay parameter value in $\{1e^{-4}, 5e^{-4}, 1e^{-3}, 5e^{-3}, 1e^{-2},5e^{-2}\}$. For ImageNet, since we use cosine learning rate schedule, for SGDM we grid search the final learning rate in $\{1e^{-3}, 5e^{-4}, 1e^{-4}, 5e^{-5}, 1e^{-5}\}$ and for the adaptive gradient methods we search the final learning rate in $\{1e^{-5}, 5e^{-6}, 1e^{-6}, 5e^{-7}, 1e^{-7}\}$.

\paragraph{LSTM for Language Modeling}
For SGDM, we grid search the learning rate in $\{100,50,30,10,1 0.1\}$ and momentum parameter between $0.5$ and $0.9$ with stepsize $0.1$ . For all the adaptive gradient methods, we fix $\beta_2=0.999$ and search $\beta_1$ between $0.5$ and $0.9$ with stepsize $0.1$, the learning rate between $0.1$ and $1e^{-5}$ in the log-grid, $\epsilon$ between $1e^{-5}$ and $1e^{-16}$ in the log-grid. For all the optimizers, we fix the weight decay parameter value as $1.2e^{-4}$ following~\citet{zhuang2020adabelief}. 

\paragraph{Transformer for Neural Machine Translation}
For SGDM, we search learning rate between $0.1$ and $10^{-5}$ in the log-grid and momentum parameter between $0.5$ and $0.9$ with stepsize 0.1. For adaptive gradient methods, we fix $\beta_1=0.9$, grid search $\beta_2$ in $\{0.98,0.99,0.999\}$, learning rate in $\{1e^{-4}, 5e^{-4}, 1e^{-3}, 1.5e^{-3}, 2e^{-3}, 3e^{-3}\}$, and $\epsilon$ between $1e^{-5}$ and $1e^{-16}$ in the log-grid. For all the optimizers we grid search weight decay parameter in $\{1e^{-4}, 5e^{-4}, 1e^{-3}, 5e^{-3}, 1e^{-2},5e^{-2}\}$. 

\paragraph{Generative Adversarial Network}
For SGDM we search the momentum parameter between $0.5$ and $0.9$ with stepsize $0.1$. For all the adaptive gradient optimziers we set $\beta_1=0.5$, search $\beta_2$ and $\epsilon$ using the same schedule as previsou subsection.

All the experiments reported are trained on NVIDIA Tesla V100 GPUs. We provide some additional information concerning the empirical experiments  for completeness.
\subsection{Image classification}
\begin{table}[tb]
  \caption{Well tuned hyperparameter configuration of the adaptive gradient methods for CNNs on CIFAR-10.}
  \label{table: hyper-cifar}
  \centering
  \scalebox{0.9}{
  \begin{tabular}{c|ccccccc}
  \toprule[1pt]
   Algorithm&  Adam & AdamW & Yogi & AdaBound & RAdam & AdaBelief & AdaMomentum \\ \midrule
   Stepsize $\alpha$& $0.001$ & $0.001$ & $0.001$ & $0.001$ & $0.001$ & $0.001$ & $0.001$ \\ \hline
   $\beta_1$& $0.9$ & $0.9$ & $0.9$ & $0.9$ & $0.9$ & $0.9$ & $0.9$  \\ \hline
   $\beta_2$& $0.999$ & $0.999$ & $0.999$ & $0.999$ & $0.999$ & $0.999$ & $0.999$ \\ \hline
   Weight decay& $5\times 10^{-4}$ & $5\times 10^{-4}$ & $5\times 10^{-4}$ & $5\times 10^{-4}$ & $5\times 10^{-4}$ & $5\times 10^{-4}$ & $5\times 10^{-4}$ \\ \hline
   $\epsilon$& $10^{-8}$ & $10^{-8}$ & $10^{-8}$ & $10^{-8}$ & $10^{-8}$ & $10^{-8}$ & $10^{-8}$ \\
   \bottomrule[1pt]
  \end{tabular}
  }
\end{table}

\paragraph{CIFAR datasets} The values of the hyperparameters after careful tuning of the reported results of the adaptive gradient methods on CIFAR-10 in the main paper is summarized in Table~\ref{table: hyper-cifar}. For SGDM, the optimal hyperparameter setting is: the learning rate is $0.1$, the momentum parameter is $0.9$, the weight decay parameter is $5 \times 10^{-4}$. For Adabound, the final learning rate is set as $0.1$ (matching SGDM) and the value of the hyperparameter gamma is $10^{-3}$.

\paragraph{ImageNet}  For SGDM, the tuned stepsize is $0.1$, the tuned momentum parameter is $0.9$ and the tuned weight decay is $1\times 10^{-4}$. For Adam, the learning rate is $0.001$, $\epsilon=1e^{-8}$, and the weight decay parameter is $1e^{-4}$. For AdaMomentum, the learning rate is $0.001$, $\epsilon=1e^{-16}$ and the weight decay parameter is $5e^{-2}$.

\subsection{LSTM on language modeling}

\begin{figure}[tb]
    \centering
    \subfigure[1-Layer LSTM.]{
      \includegraphics[width=.33\linewidth]{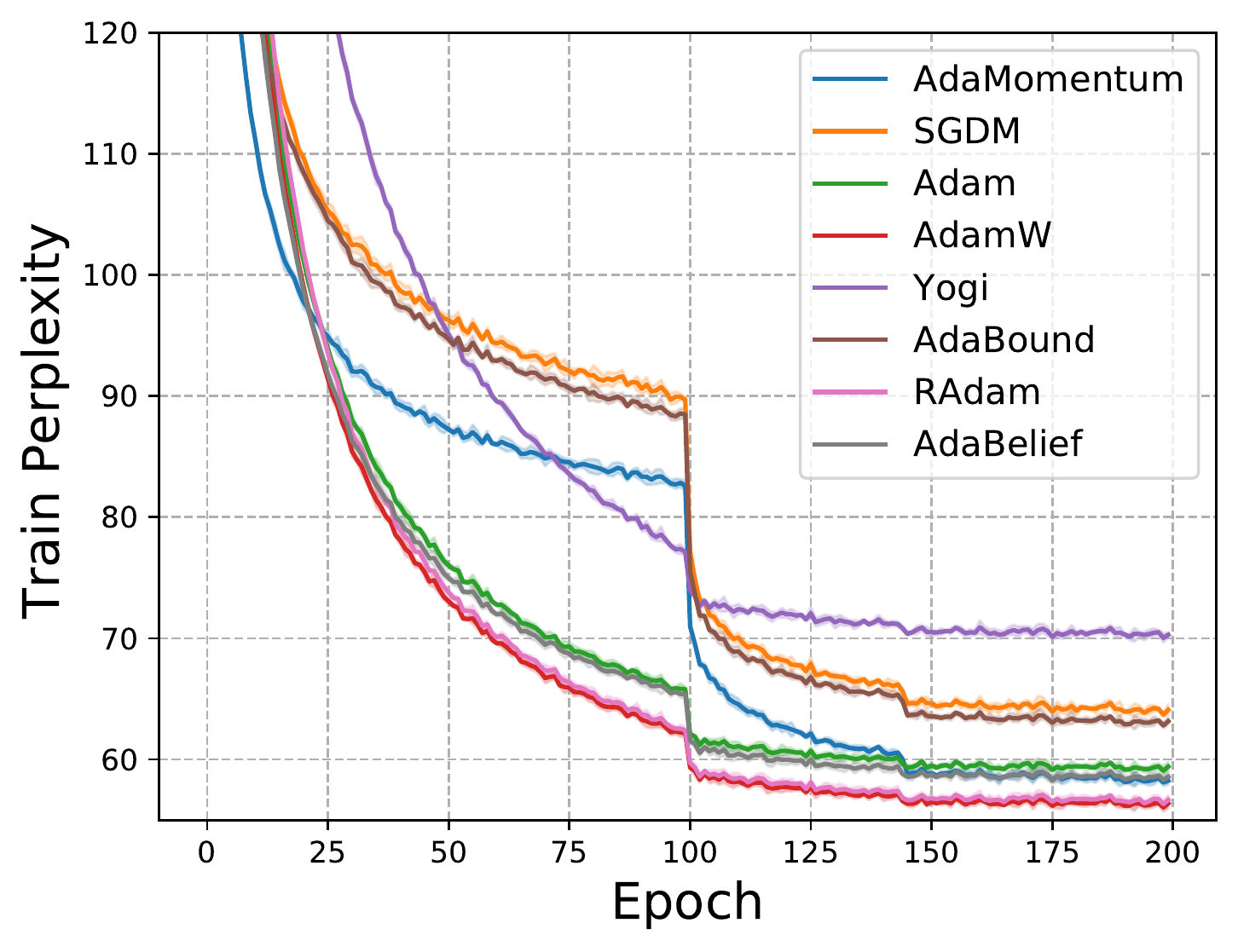}  
    }\hspace{-10mm}
    \hfill
    \subfigure[2-Layer LSTM.]{
      \includegraphics[width=.33\linewidth]{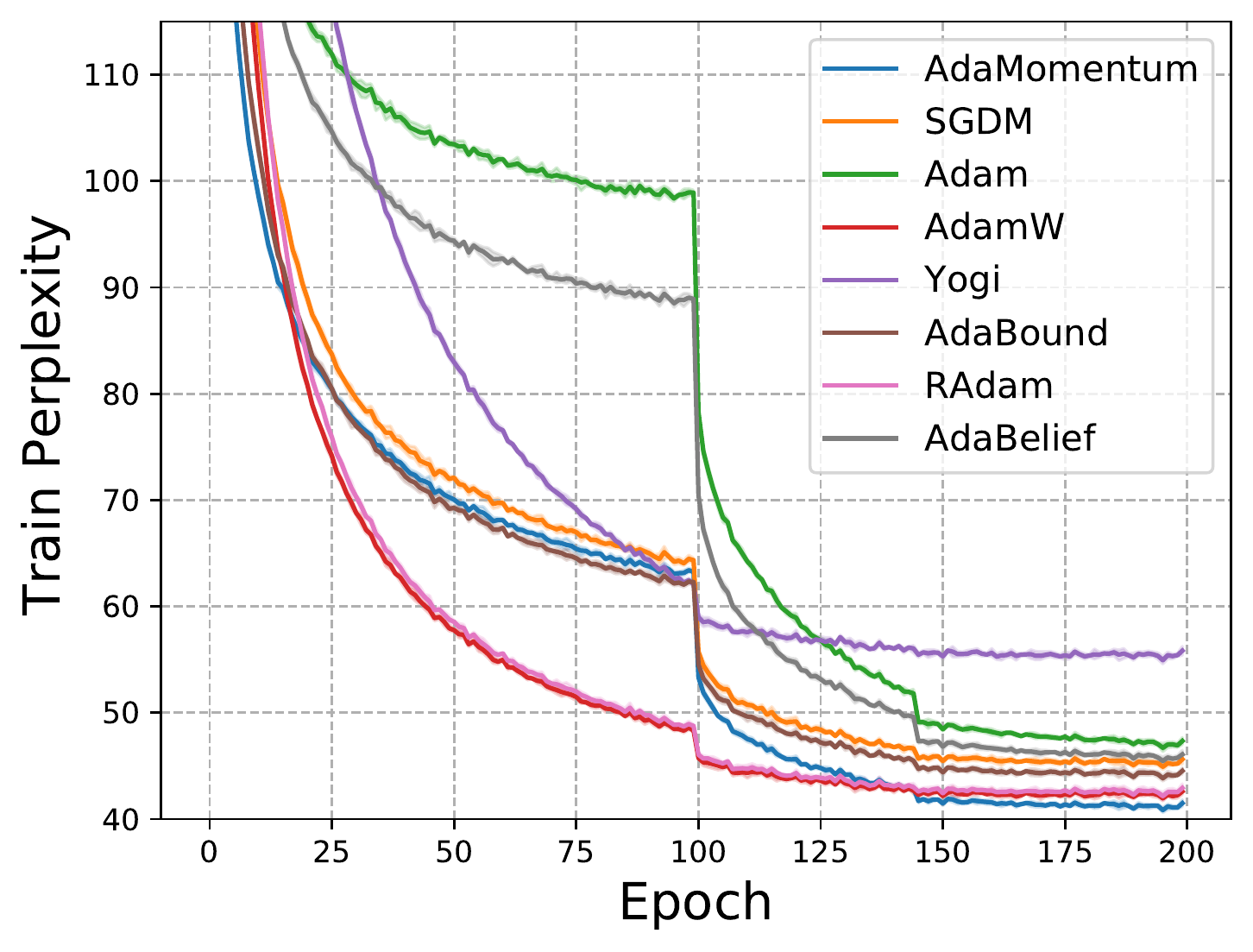}  
    }\hspace{-10mm}
    \hfill
    \subfigure[3-Layer LSTM.]{
      \includegraphics[width=.33\linewidth]{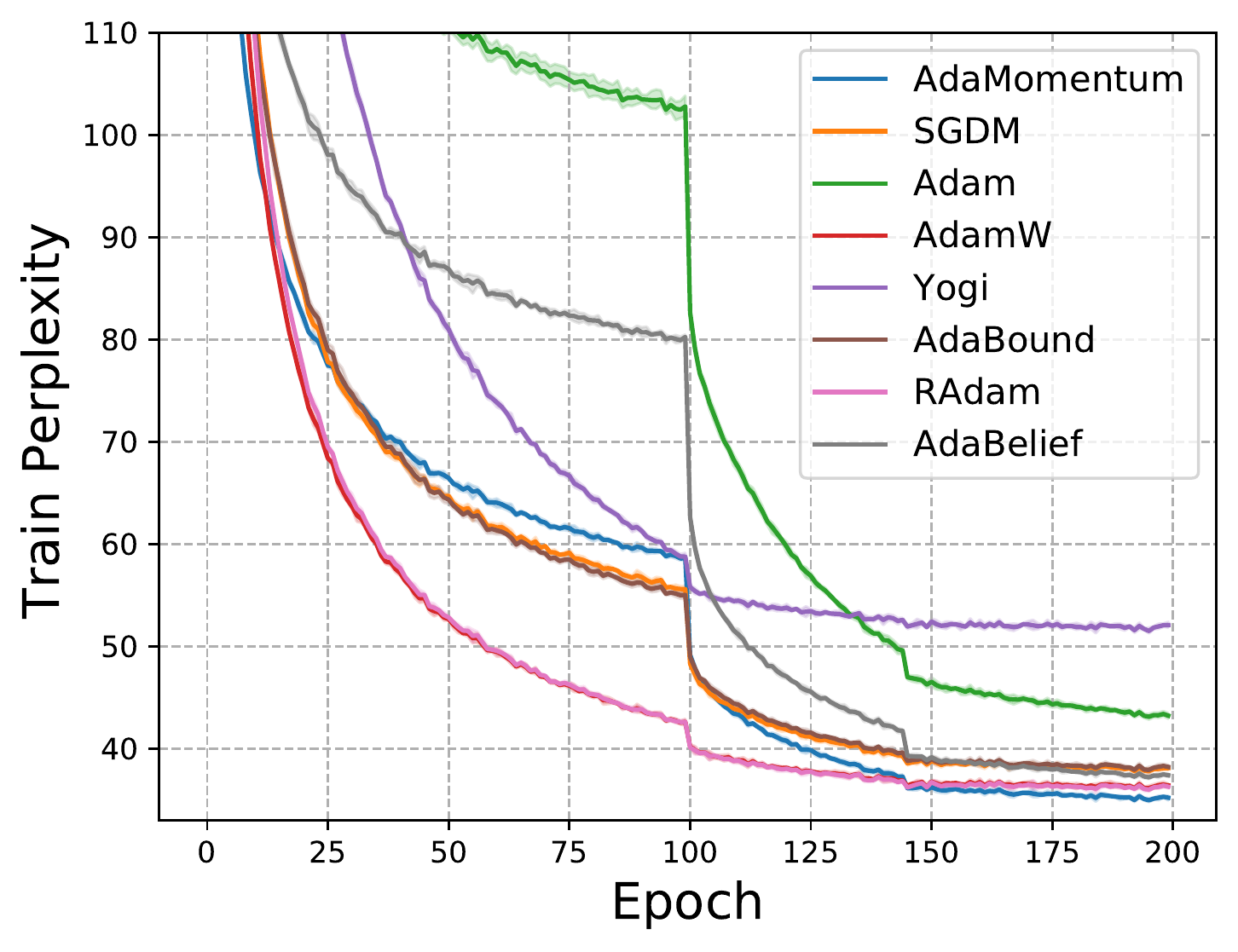}  
    }
    \caption{Train perplexity curve on Penn Treebank~\cite{marcus-etal-1993-building} dataset.}
    \label{fig: lstm-train}
\end{figure}
\begin{figure*}[tb]
  \centering
  \subfigure[1-Layer LSTM.]{
    \includegraphics[width=.33\linewidth]{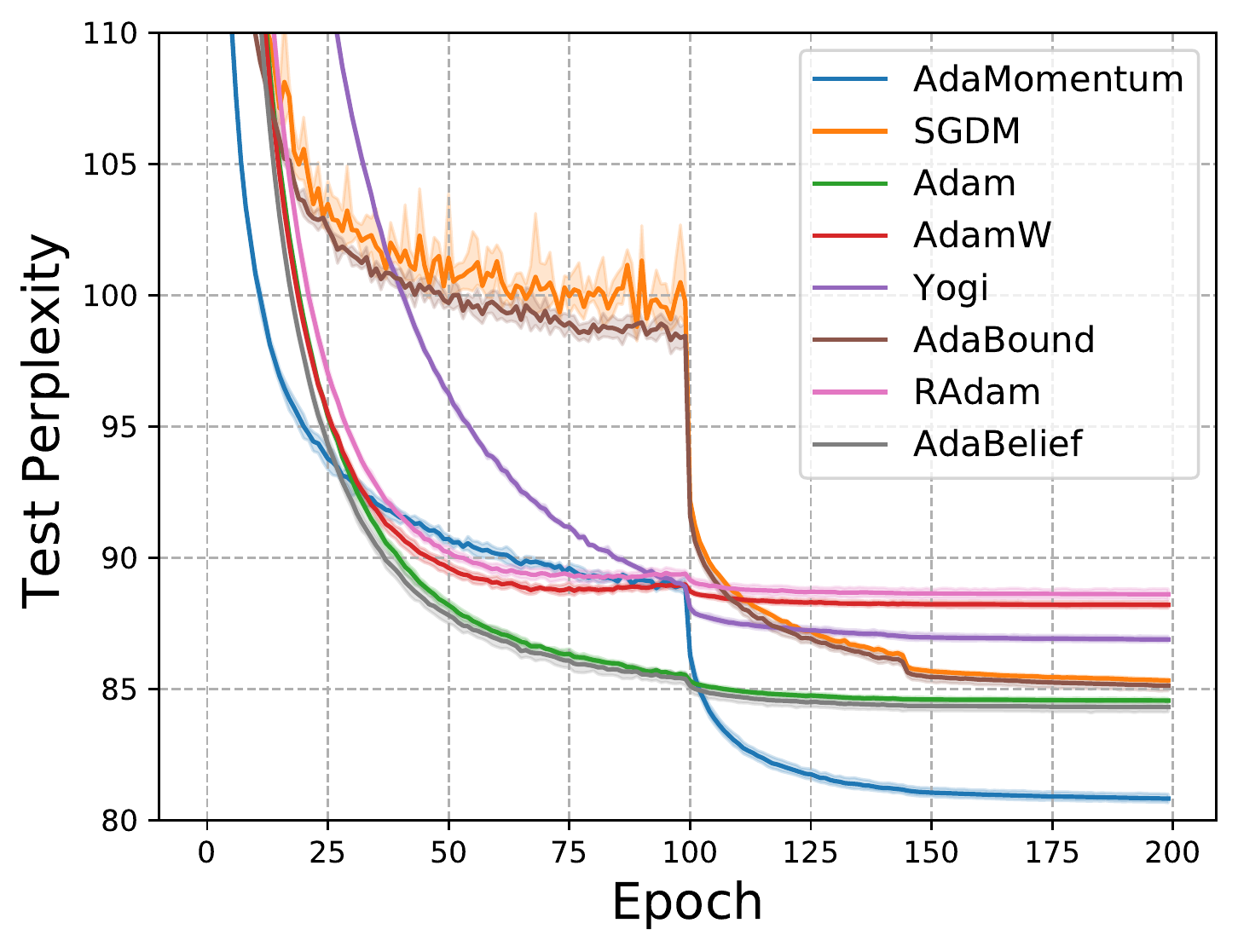}  
  }\hspace{-10mm}
  \hfill
  \subfigure[2-Layer LSTM.]{
    \includegraphics[width=.33\linewidth]{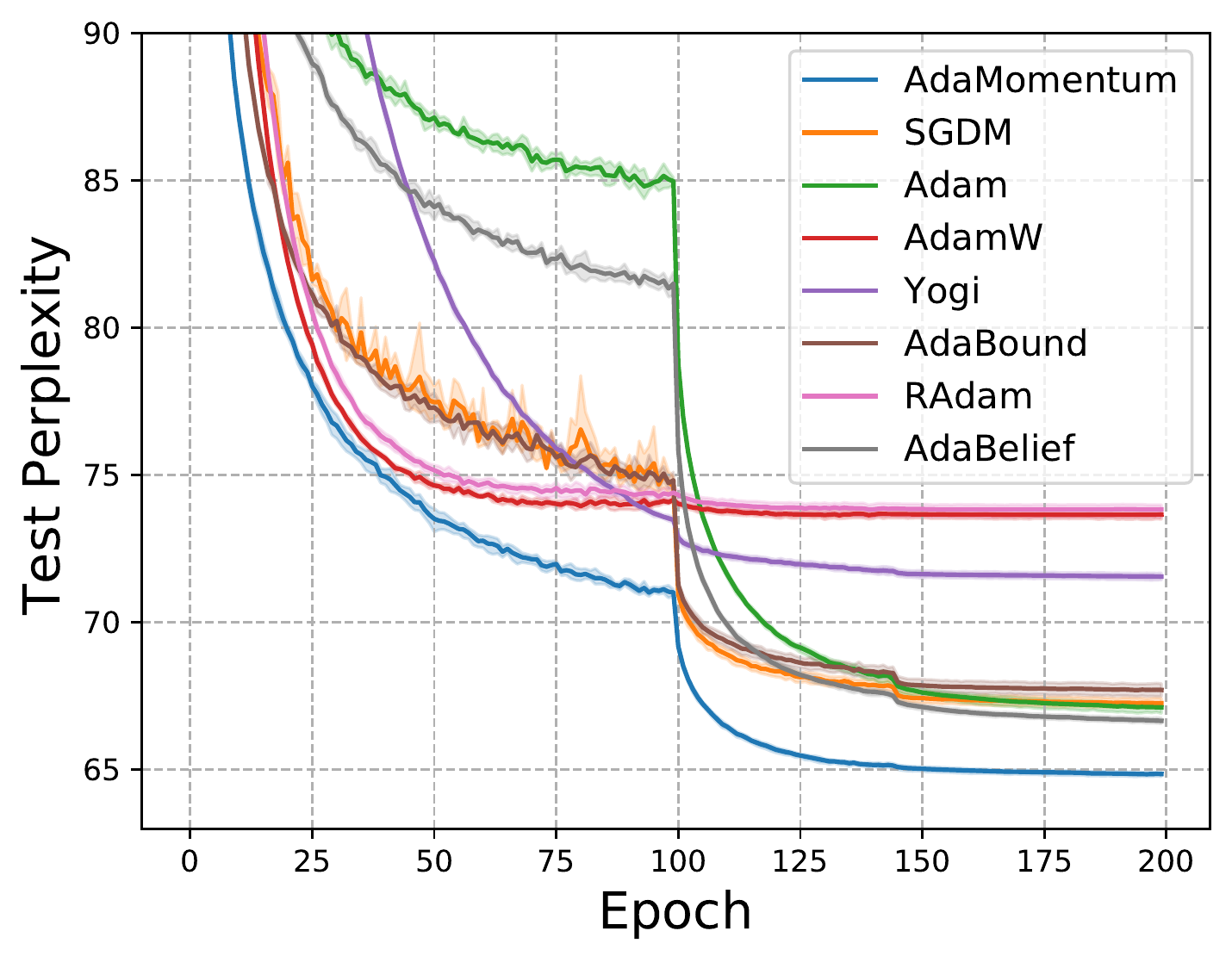}  
  }\hspace{-10mm}
  \hfill
  \subfigure[3-Layer LSTM.]{
    \includegraphics[width=.33\linewidth]{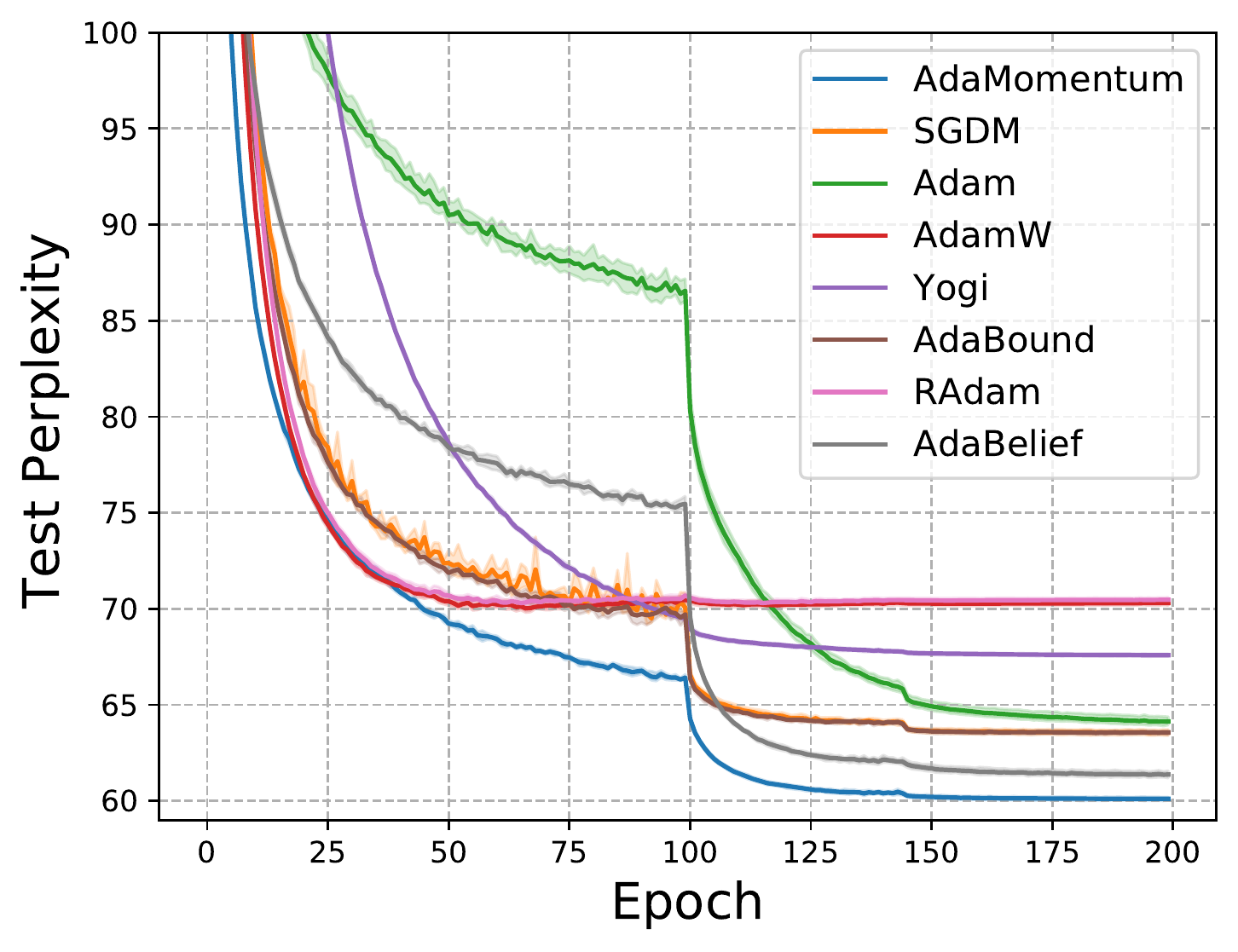}  
  }
  \caption{Test perplexity curve on Penn Treebank~\cite{marcus-etal-1993-building} dataset.}
  \label{fig: lstm-test}
\end{figure*}
\begin{table}[tb]
    \caption{Well tuned hyperparameter configuration of adaptive gradient methods for 1-layer-LSTM on Penn Treebank dataset.}
    \label{table: hyper-lstm-1}
    \centering
    \scalebox{0.85}{
    \begin{tabular}{c|ccccccc}
    \toprule[1pt]
     Algorithm& Adam & AdamW & Yogi & AdaBound & RAdam & AdaBelief & AdaMomentum \\ \midrule
     Stepsize $\alpha$& $0.001$ & $0.001$ & $0.01$ & $0.01$ & $0.001$ & $0.001$ & $0.001$ \\ \hline
     $\beta_1$& $0.9$ & $0.9$ & $0.9$ & $0.9$ & $0.9$ & $0.9$ & $0.9$  \\ \hline
     $\beta_2$& $0.999$ & $0.999$ & $0.999$ & $0.999$ & $0.999$ & $0.999$ & $0.999$ \\ \hline
     Weight decay& $1.2\times 10^{-4}$  & $1.2\times 10^{-4}$  & $1.2\times 10^{-4}$  & $1.2\times 10^{-4}$  & $1.2\times 10^{-4}$  & $1.2\times 10^{-4}$ & $1.2\times 10^{-4}$  \\ \hline
     $\epsilon$& $10^{-12}$ & $10^{-12}$ & $10^{-8}$ & $10^{-8}$ & $10^{-12}$ & $10^{-16}$ & $10^{-16}$ \\
     \bottomrule[1pt]
    \end{tabular}
    }
\end{table}
\begin{table}[tb]
    \caption{Well tuned hyperparameter configuration of adaptive gradient methods for 2-layer-LSTM on Penn Treebank dataset.}
    \label{table: hyper-lstm-2}
    \centering
    \scalebox{0.85}{
    \begin{tabular}{c|ccccccc}
    \toprule[1pt]
     Algorithm& Adam & AdamW & Yogi & AdaBound & RAdam & AdaBelief & AdaMomentum \\ \midrule
     Stepsize $\alpha$& $0.01$ & $0.001$ & $0.01$ & $0.01$ & $0.001$ & $0.01$ & $0.001$ \\ \hline
     $\beta_1$& $0.9$ & $0.9$ & $0.9$ & $0.9$ & $0.9$ & $0.9$ & $0.9$  \\ \hline
     $\beta_2$& $0.999$ & $0.999$ & $0.999$ & $0.999$ & $0.999$ & $0.999$ & $0.999$ \\ \hline
     Weight decay& $1.2\times 10^{-4}$  & $1.2\times 10^{-4}$  & $1.2\times 10^{-4}$  & $1.2\times 10^{-4}$  & $1.2\times 10^{-4}$  & $1.2\times 10^{-4}$ & $1.2\times 10^{-4}$  \\ \hline
     $\epsilon$& $10^{-12}$ & $10^{-12}$ & $10^{-8}$ & $10^{-8}$ & $10^{-12}$ & $10^{-12}$ & $10^{-16}$ \\
     \bottomrule[1pt]
    \end{tabular}
    }
\end{table}
\begin{table}[tb]
    \caption{Well tuned hyperparameter configuration of adaptive gradient methods for 3-layer-LSTM on Penn Treebank dataset.}
    \label{table: hyper-lstm-3}
    \centering
    \scalebox{0.85}{
    \begin{tabular}{c|ccccccc}
    \toprule[1pt]
     Algorithm& Adam & AdamW & Yogi & AdaBound & RAdam & AdaBelief & AdaMomentum \\ \midrule
     Stepsize $\alpha$& $0.01$ & $0.001$ & $0.01$ & $0.01$ & $0.001$ & $0.01$ & $0.001$ \\ \hline
     $\beta_1$& $0.9$ & $0.9$ & $0.9$ & $0.9$ & $0.9$ & $0.9$ & $0.9$  \\ \hline
     $\beta_2$& $0.999$ & $0.999$ & $0.999$ & $0.999$ & $0.999$ & $0.999$ & $0.999$ \\ \hline
     Weight decay& $1.2\times 10^{-4}$  & $1.2\times 10^{-4}$  & $1.2\times 10^{-4}$  & $1.2\times 10^{-4}$  & $1.2\times 10^{-4}$  & $1.2\times 10^{-4}$ & $1.2\times 10^{-4}$  \\ \hline
     $\epsilon$& $10^{-12}$ & $10^{-12}$ & $10^{-8}$& $10^{-8}$ & $10^{-12}$ & $10^{-12}$ & $10^{-16}$ \\
     \bottomrule[1pt]
    \end{tabular}
    }
\end{table}

The training and testing perplexity curves are illustrated in Figure~\ref{fig: lstm-train} and~\ref{fig: lstm-test}. We can clearly see that AdaMomentum is able to make the perplexity descent faster than SGDM and most other adaptive gradient methods during training and mean while generalize much better in testing phase. In experimental settings, the size of the word embeddings is $400$ and the number of hidden units per layer is $1150$. We employ dropout in training and the dropout rate for RNN layers is $0.25$ and the dropout rate for input embedding layers is $0.4$.

The optimal hyperparameters of adaptive gradient methods for 1-layer, 2-layer and 3-layer LSTM are listed in Tables~\ref{table: hyper-lstm-1}, \ref{table: hyper-lstm-2} and~\ref{table: hyper-lstm-3} respectively. For SGDM, the Well tuned stepsize is $30.0$ and the momentum parameter is $0.9$. For Adabound, the final learning rate is set as $30.0$ (matching SGDM) and the value of the hyperparameter gamma is $10^{-3}$. 

\subsection{Transformer on neural machine translation}
\begin{table}[tb]
    \caption{Well tuned hyperparameter configuration of adaptive gradient methods for transformer on IWSTL'14 DE-EN dataset.}
    \label{table: transformer}
    \centering
    \begin{tabular}{c|cccc}
    \toprule[1pt]
     Algorithm& Adam & AdamW & AdaBelief & AdaMomentum \\ \midrule
     Stepsize $\alpha$& $0.0015$ & $0.0015$ & $0.0015$ & $0.0005$ \\ \hline
     $\beta_1$& $0.9$ & $0.9$ & $0.9$ & $0.9$  \\ \hline
     $\beta_2$& $0.98$ & $0.98$ & $0.999$ & $0.999$  \\ \hline
     Weight decay& $10^{-4}$  & $10^{-4}$  & $ 10^{-4}$  & $10^{-4}$  \\ \hline
     $\epsilon$& $10^{-8}$ & $10^{-8}$ & $10^{-16}$& $10^{-16}$  \\
     \bottomrule[1pt]
    \end{tabular}
\end{table}
For transformer on NMT task, the well tuned hyperparameter values are summarized in Table~\ref{table: transformer}. The stepsize of SGDM is $0.1$ and the momentum parameter of SGDM is $0.9$. Initial learning rate is $10^{-7}$ and the minimum learning rate threshold is set as $10^{-9}$ in the warm-up process for all the optimizers. 

\subsection{Generative Adversarial Network}

\begin{table}[tb]
  \caption{Well tuned hyperparameter configuration of adaptive gradient methods for BigGAN with consistency regularization.}
  \label{table: GAN}
  \centering
  \scalebox{0.85}{
  \begin{tabular}{c|cccccc}
  \toprule[1pt]
   Algorithm& Adam & Yogi & AdaBound & RAdam & AdaBelief & AdaMomentum \\ 
   \midrule
   $\beta_1$& $0.5$ & $0.5$ & $0.5$ & $0.5$ & $0.5$ & $0.5$  \\ \hline
   $\beta_2$& $0.999$ & $0.999$ & $0.999$ & $0.999$ & $0.999$ & $0.999$  \\ \hline
   Weight decay& $0 $  & $0$  & $0$  & $0$  & $0$  & $0$  \\ \hline
   $\epsilon$& $10^{-8}$ & $10^{-8}$ & $10^{-8}$& $10^{-8}$ & $10^{-16}$ & $10^{-16}$ \\
   \bottomrule[1pt]
  \end{tabular}
  }
\end{table}

\begin{figure}[tb]
  \centering
  \subfigure[DCGAN trained using random seed 0 ( best FID score 43.52, iteration 26000).]{
    \includegraphics[width=.45\linewidth]{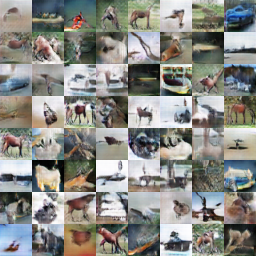}  
  }\hspace{-10mm}
  \hfill
  \subfigure[BigGAN trained using random seed 2 ( best FID score 7.07, iteration 92000).]{
    \includegraphics[width=.45\linewidth]{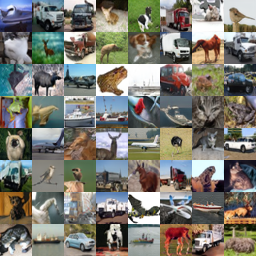}  
  }
  \caption{Generated figures trained on CIFAR-10 optimizing with AdaMomentum.}
  \label{fig: gan-visualize}
\end{figure}

The optimal momentum parameters of SGD for all GANs are $0.9$. For adaptive gradient methods, the well tuned hyperparameter values for BigGAN with consistency regularization are summarized in Table~\ref{table: GAN}. We implement the GAN experiments adapting the code from public repository~\footnote{\url{https://github.com/POSTECH-CVLab/PyTorch-StudioGAN}}. We sample two visualization results of generated samples of GAN training with AdaMomentum in Figure~\ref{fig: gan-visualize}.